\documentclass[11pt]{article}

\usepackage{amsmath,amsfonts,bm}

\def\eqref#1{equation~\ref{#1}}

\def\martin{\nu}

\DeclareMathAlphabet{\mathsfit}{\encodingdefault}{\sfdefault}{m}{sl}
\SetMathAlphabet{\mathsfit}{bold}{\encodingdefault}{\sfdefault}{bx}{n}

\usepackage{palatino}
\usepackage[letterpaper,margin=1in]{geometry}

\usepackage[utf8]{inputenc} 
\usepackage[T1]{fontenc}    
\usepackage{hyperref}       
\usepackage{amsfonts,amsmath,amssymb,amsthm}       
\usepackage{array}
\usepackage{url}            
\usepackage{booktabs}       
\usepackage{nicefrac}       
\usepackage{microtype}      
\usepackage[dvipsnames]{xcolor}         
\usepackage{bm}
\usepackage{algorithm}

\usepackage{natbib}
\usepackage{graphicx}
\usepackage{subcaption}
\usepackage{hyperref}

\usepackage{bigints}
\usepackage{amssymb,amsopn,algorithm,algorithmic,float,bbm,bm,enumerate,color,multirow,gensymb}
\usepackage{comment}
\usepackage{afterpage}
\usepackage{thmtools,thm-restate}

\newcommand{\beq}{\begin{equation}}
\newcommand{\eeq}{\end{equation}}

\newcommand{\norm}[1]{\Vert #1 \Vert}

\newcommand\E{\mathbb{E}}
\renewcommand\P{\mathbb{P}}

\newcommand\R{\mathbb{R}}

\newcommand{\x}{\mathbf{x}}

\newcommand{\cC}{{\cal C}}

\newcommand{\cL}{{\cal L}}
\newcommand{\cM}{{\cal M}}

\newcommand{\cD}{{\cal D}}

\newcommand{\hH}{\hat{H}}

\newcommand{\hV}{\hat{V}_t^{-1/2}}
\newcommand{\barm}{\bar{m}}
\newcommand{\barv}{\bar{v}}
\newcommand{\etat}{\eta}
\newcommand{\kV}{\hat{V}_{t-1}^{-1/2}}

\newcommand{\lr}{\kappa}
\newcommand{\sumL}{\hat{L}}

\newtheorem{remark}{Remark}[section]

\def\martin{\nu}
\def\localnoise{\Delta}
\def\stocdelta{\delta_c}

\newcommand {\commentout}[1] {}

\def\ints{{{\rm Z} \kern -.35em {\rm Z} }}  
\def\smallints{{{\rm Z} \kern -.3em {\rm Z} }}  
\def\pints{{{\rm I} \kern -.15em {\rm N} }}      
\newcommand{\reals}{\mathbb R}

\def\cplx{{{\rm I} \kern -.45em {\rm C} }}       
\def\l2{\rm {\mathcal L}^{2}(\reals)}            

\newtheorem{nad}{Notation and Definitions}[section]

\newtheorem{theorem}{Theorem}[section]
\newtheorem{lemma}[theorem]{Lemma}

\newtheorem{corollary}{Corollary}

\newtheorem{assumption}{Assumption}
\newtheorem{property}{Property}

\newcommand{\be}{\begin{eqnarray}}
\newcommand{\ee}{\end{eqnarray}}
\newcommand{\bea}{\begin{eqnarray}}
\newcommand{\eea}{\end{eqnarray}}
\newcommand{\beaa}{\begin{eqnarray*}}
\newcommand{\eeaa}{\end{eqnarray*}}
\newcommand{\bnad}{\begin{nad}}
\newcommand{\enad}{\end{nad}}

\newcommand{\logcd}{\frac{\log^{1.5}(Cd/\delta)}{\sqrt{b}}}
\newcommand{\logctdsq}{\log^3(CT d^2 /\delta)}
\newcommand{\logcsqdt}{\frac{\log^{1.5}(Ctd^2/\delta)}{\sqrt{b}}}
\newcommand{\logd}{\frac{\log^{1.5}(d/\delta)}{\sqrt{b}}}
\newcommand{\logtd}{\frac{\log^{1.5}(CKTd/\delta)}{\sqrt{b}}}
\newcommand{\logstd}{\frac{\log^{1.5}(CKtd/\delta)}{\sqrt{b}}}

\newcommand{\logdsqt}{\frac{\log^{1.5}(CKtd^2/\delta)}{\sqrt{b}}}
\newcommand{\logtdsq}{\frac{\log^{1.5}(CKd^2T^2/\delta)}{\sqrt{b}}}
\newcommand{\logtsqd}{\frac{\log^{1.5}(CKdT^2/\delta)}{\sqrt{b}}}

\newcommand{\sumv}[1]{\frac{1}{{#1}} \sum_{c=1}^{#1}}

\newcommand{\sk}{\texttt{sk}}
\newcommand{\desk}{\texttt{desk}}
\newcommand{\comp}{\texttt{sk}}
\newcommand{\decomp}{\texttt{desk}}

\newcommand{\eps}{\varepsilon}

\newcommand{\calQ}{{\cal Q}}

\newcommand{\bscale}{b_0}

\newcommand{\intdim}{\mathcal{I}}
\newcommand{\localgradbnd}{2\sqrt{L} \sqrt{\cL(x_t)} + 2\sqrt{2\localnoise^2 \ln \frac{2}{\stocdelta}} + \localnoise}

\theoremstyle{definition}
\newtheorem{definition}[theorem]{Definition}

\usepackage[dvipsnames]{xcolor}
\usepackage[]{color-edits}
\addauthor{ab}{violet}

\date{}

\title{Sketched Adaptive Federated Deep Learning:\\ A Sharp Convergence Analysis}

\author{
Zhijie Chen \\
Siebel School of Computing and Data Science\\
University of Illinois at Urbana-Champaign\\
\texttt{lucmon@illinois.edu}
\and
Qiaobo Li \\
Siebel School of Computing and Data Science\\
University of Illinois at Urbana-Champaign\\
\texttt{qiaobol2@illinois.edu}
\and
Arindam Banerjee \\
Siebel School of Computing and Data Science\\
University of Illinois at Urbana-Champaign\\
\texttt{arindamb@illinois.edu}}

\begin{document}

\maketitle

\begin{abstract}
  Combining gradient compression methods and adaptive optimizers is a desirable goal in federated learning (FL), with potential benefits on both fewer communication rounds and less per-round communication. In spite of the preliminary empirical success of compressed adaptive methods, existing convergence analyses show the communication cost to have an effectively linear dependence on the number of parameters, which is prohibitively high for modern deep learning models.

In this work, we introduce specific sketched adaptive federated learning (SAFL) algorithms and, as our main contribution, provide theoretical convergence analyses with guarantees on communication cost depending only logarithmically on the number of parameters. Unlike existing analyses, we show that the entry-wise sketching noise existent in the preconditioners and the first moments of SAFL can be implicitly addressed by leveraging the intrinsic dimension of loss Hessian, which is reckoned significantly smaller than the full dimensionality in deep learning models.  Our theoretical claims are supported by empirical studies on vision and language tasks, and in both supervised fine-tuning and training-from-scratch regimes. Surprisingly, as a by-product of our analysis, the proposed SAFL methods are competitive with the state-of-the-art communication-efficient federated learning algorithms based on error feedback.

\end{abstract}

\section{Introduction}
Despite the recent success of federated learning (FL), the cost of communication arguably remains the main challenge. \cite{wang2023cocktailsgd} showed that a 20 Gbps network bandwidth is necessary to  bring the communication overhead to a suitable scale for finetuning GPT-J-6B, which is unrealistic in distributed settings. Even with good network conditions, reduction of communication complexity means one can train much larger models given the same communication budget. 

The communication cost of vanilla FL can be represented as $O(dT)$, where $d$ is the ambient dimension of the parameter space, i.e. the number of parameters, and $T$ is the number of communication rounds for convergence. Various methods have been proposed to minimize $T$, e.g., local training~\citep{stich2018local}, large batch training~\citep{xu2023slamb}. 
Folklores in centralized training regimes suggest that $T$ heavily relies on the choice of optimizers, where adaptive methods usually demonstrate faster convergence and better generalization performance, especially in transformer-based machine learning models~\citep{reddi2019convergence}. 



The alternative approach of reducing communication costs is to be more thrifty on the communication bits at a single round, i.e., to reduce the $O(d)$ factor, which is dominant in the communication complexity for modern neural networks where $d \gg T$,  to $O(b)$. Considerable efforts have been devoted to design efficient gradient compression methods, which compress a vector of dimension $d$ to an effective size $b$.  Popular gradient compression methods include quantization~\citep{alistarh2017qsgd, chen2023nqfl, reisizadeh2020fedpaq, liu2023communication}, sparsification ~\citep{alistarh2018convergence, wu2018error, rothchild2020fetchsgd} and sketching~\citep{spring2019compressing, jiang2024correlation,song2023sketching}. 

Denote $\cC$ as the compression operator over vector $x$. The compression error $\omega$ can be characterized by $\Vert \cC(x) - x \Vert \le \omega \Vert x \Vert$. The convergence rates of such compressed gradient methods heavily depend on $\omega$. For the family of unbiased compressors, $\omega$ can have linear dependence on $d$. For instance, $l_2$-quantization and unbiased RandK sparsifier~\citep{beznosikov2023biased} achieves $\omega = \frac{d}{b}-1$, and PermK~\citep{szlendak2021permutation}, which is a statistically dependent variant of RandK in the FL setting, achieves a constant level $\omega$ only when the sketch size is proportional to $d$.  Recent works show that the convergence rate depends on the number of clients $C$ being involved in each round. For instance, PermK~\citep{szlendak2021permutation} achieves an $O(\frac{d-1}{C-1})$ compression error when $C \ge d$. While an exciting advance, arguably in many FL settings with modern deep learning models, the number of parameters $d$ (hundreds of billions or more) is much larger than the number of clients $C$  (millions). The difference in magnitude makes the compensation of dimension hardly achievable in practice. The usage of such unbiased compressors effectively leads to dimension-dependent convergence rate in in compressed gradient based FL methods such as MARINA~\citep{gorbunov2021marina} and MARINA-P~\citep{sokolov2024marina}.

Biased gradient compressors are capable of achieving significantly lower compression error than the unbiased counterparts. TopK and biased RandK, which are commonly-used contractive compressors, achieve $\omega \le (1-\frac{b}{d})$. The issue of the biased methods in leading to divergence under even simple cases~\citep{beznosikov2023biased}  can be mitigated by introducing error feedback (EF) mechanisms~\citep{seide20141}, and the theoretical guarantees are provided in~\citep{stich2018local}.
However, the state-of-the-art error feedback EF21~\citep{richtarik2021ef21} utilizing the Markov compressor, and its subsequent variants~\citep{richtarik2024error, fatkhullin2024momentum} still suffer from the \textit{distortion error} which is proportional to $\frac{d}{b}$. The dimensional dependence is inherited to the convergence rate of CocktailSGD~\citep{wang2023cocktailsgd}, and 3PC~\citep{richtarik20223pc} that employ biased gradient compressions. Furthermore, most of the developments on EF do not explicitly show compatibility with {\em adaptive methods, which involve anisotropic and nonlinear updates}~\citep{tang20211}. Indeed, the design and analysis of {\bf communication-efficient adaptive FL algorithms} pose non-trivial challenges.

These existing works on the theory of communication-efficient adaptive FL algorithms have arguably alarming results, which do not match practice. 
The existing analyses show that the iterations $T$ needed for convergence can be inversely proportional to the compression rate~\citep{chen2022efficient, song2023sketching}. For constant per-round communication bits, the bounds indicate the iteration complexity to scale as $O(d)$, i.e., linearly with the ambient dimensionality, which is prohibitively large for modern deep learning models. The mismatch between such potential theory issues vs.~preliminary empirical promise 
has prevented wide adoption of such adaptive FL algorithms. 

Furthermore, the involvement of gradient compression calls for designing adequate transmission mechanisms. For sparsifying compressions, such as TopK and RandK, the average of sparse client gradients is possibly dense, which increases the downlink (server-to-client) transmission overhead. In the worst case, a plain average of the client gradients in MARINA~\citep{gorbunov2021marina} leads to $bC$ in the number of non-zero bits. 
FetchSGD~\citep{rothchild2020fetchsgd} mitigates the problem by adopting an extra call of topK compressor on the server side at additional compression costs.



In this work, we first introduce a family of Sketched Adaptive FL (SAFL) algorithms, with flexibility on the choice of sketching methods and adaptive optimizers, that simultaneously guarantees convergence and reduces per round bits towards improved  communication efficiency. At a high level, SAFL algorithms are analogous to previous attempts~\citep{tang20211,chen2022efficient, wang2022communication}, which showed preliminary empirical success of applying gradient compression with adaptive optimizers in FL. Our SAFL algorithms adopt {\em unbiased gradient compressors} based on random linear sketching and hence {\em eliminates the need for error feedbacks}. The linearity of gradient compressions in SAFL also avoids an extra round of server side compression required in sparsification~\citep{stich2018sparsified} and quantization~\citep{reisizadeh2020fedpaq}.


As a {\bf major contribution of our current work}, we provide convergence rates of the proposed SAFL algorithms 
that depends only logarithmically (instead of linearly) on the ambient dimension $d$. The central technical challenge in addressing the dimensional dependence is to handle the entry-wise sketching noise in both the preconditioners and the first moments of the adaptive optimizers, which has been acknowledged non-trivial~\citep{tang20211, wang2022communication}. 
Our sharper analysis is built based on the intrinsic dimension (instead of the ambient dimension $d$) of the loss Hessian in deep learning, i.e., the ratio of sum of absolute eigenvalues over the largest eigenvalue. Recent observations on the Hessian spectrum of deep learning models have demonstrated that the intrinsic dimension is significantly smaller than the ambient dimension, by showing the eigenvalues decay sharply, with most eigenvalues being close to zero~\citep{ghorbani2019investigation, zhang2020adaptive, li2020hessian, liao2021hessian, liu2023sophia}, and even arguably conforming with a power-law decay~\citep{xie2022power,zhang2024transformers}. In contrast, the conventional smoothness conditions assume uniform curvature in all directions which can be overly pessimistic in the context of deep learning. 
This specific eigenspectrum structure provides significant advantages in the sharp analysis of sketching noise in adaptive methods. The SAFL algorithms do not involve computing the Hessian eigenspectrum, which is only used for the convergence analysis. Our analysis leverages the anisotropic smoothness structure, leading to the following main contributions:

(1) We introduce the sketched adaptive FL (SAFL) framework  which combines random sketching and adaptive methods. While the preconditoner in adaptive methods morphs the shape of sketching noise, posing challenges in leveraging the fast-decaying Hessian eigenstructure, we prove that the proposed sketching effectively balances iteration complexity and sketching dimension $b$.
We derive a high probability bound showing that a sketch size of $b = O(\log d)$ suffices to achieve an asymptotic $O(1/\sqrt{T})$ dimension-independent convergence rate in non-convex deep learning settings.


(2) Distinct from the existing works~\citep{reddi2020adaptive, xie2020cser}, we provide a general convergence analysis without assumptions on the gradient norm bounds on both the server and client sides. We demonstrate that although the gradient norm does not possess a uniform bound on the entire space, the proposed algorithm SAFL automatically generates bounded gradients along the entire optimization trajectory. The analysis involves a careful analysis on connecting the noisy local training steps with the global loss.


(3) We validate our theoretical claims with empirical evidence on deep learning models from vision (ResNet, Vision Transformer) and language (BERT) tasks. We cover both fine-tuning and training-from-scratch regimes. Furthermore, SAFL achieves comparable performance with the full-dimensional unsketched adaptive optimizers, and are competitive with the state-of-the-art communication-efficient FL algorithms based on error feedback and adaptive methods. 

\section{Related Works}
\textbf{Communication-efficient federated optimization.} There have been rapid advances in communication-efficient federated optimization in recent years.  Local training, i.e. running SGD independently in parallel on different clients, is the off-the-shelf training mechanism which ideally reduces the frequency of communication. \cite{stich2018local} shows local SGD achieves the same convergence rate as mini-batch SGD. \cite{wang2019adaptive} study the effect of the frequency in model averaging, and propose adaptive communication strategies. \cite{mishchenko2022proxskip} prove the local gradient step can surprisingly accelerate the training process, which offers non-trivial advantages over SGD.

Besides the advances in efficient training mechanisms, applying gradient compression is another promising research thread in communication-efficient learning. In principle, gradient compression methods reduce the communication bits per round with negligible increase of overhead in the convergence rate. Various gradient compression methods have been proposed and exhibited preliminary improvement in practice. Quantization is one of the popular compression schemes which adopts lower bits to represent data originally represented by 32 bits on each dimension. \cite{tang20211} propose 1-bit Adam based on the stability of Adam's variance term during the training time. \cite{li20221} 
improve 1bit-Adam using large batch training and adaptive layerwise learning rates. \cite{tang2024z} propose a sign-based unbiased quantization method that controls the bias of signSGD~\citep{bernstein2018signsgd} by injecting random noise prior to the compression.

Another popular gradient compression method is sparsification, where the transmitted bits solely come from the most significant values in the model update. The communication cost is proportional to the number of non-zero elements in the sparsified gradient. Deterministic sparsification methods are simpler in practice, e.g., Random-k~\citep{wangni2018gradient}, Top-k~\citep{stich2018sparsified, shi2019convergence, li2022distributed, xu2023slamb}, deep gradient compression~\citep{lin2017deep}. However, the consequential biased gradient estimation reportedly hurts training performance and leads to worse generalization ~\citep{beznosikov2023biased}. Error compensation techniques~\citep{zheng2019communication, richtarik2021ef21} are necessary to mitigate the effect. \cite{rothchild2020fetchsgd} first propose to apply error compensation on the server side to support sparse client participation. \cite{wang2021error} apply targeted error compensation to specific components of the sparsified model updates. 

Distinguished from all the methods above, sketching has gained increasing popularity because of several favorable properties including mergibility and unbiasedness. Sketching methods project the entire gradient vector into a tiny subspace. Our method also falls into this category and is well-compatible with adaptive optimizers and associated with a sharper convergence analysis. \cite{ivkin2019communication} utilize Count-Sketch~\citep{charikar2002finding} to estimate the heavy-hitters of a gradient vector. \cite{vargaftik2021drive} estimate the coordinates of a gradient with structured random rotations in a high-dimensional sphere.  \cite{rabbani2021comfetch} apply sketching to model weights to improve downlink communication efficiency.


Theoretical analysis on communication-efficient federated learning is also a central topic in this thread. \cite{ivkin2019communication} develop the convergence guarantees for Count-Sketch in a strongly-convex setting. \cite{chen2021quantized, chen2022efficient} conduct a convergence analysis for quantized Adam with error compensation. \cite{haddadpour2021federated} provide a unified convergence analysis on periodical compressed communication mechanism based on quantization and sparsification. \cite{wang2022communication} study the convergence properties of communication-efficient adaptive gradient methods under biased compressors. \cite{song2023sketching} provide the first convergence result of random sketching in the non-convex setting, but the upper bound comes with a dimension dependence.


\textbf{Noise in Deep Learning.}  In our work, we deal with noises from various sources. There have been numerous literatures discussing the noise in neural network training. However, high-probability bounds are indeed quite limited, as the mainstream of analysis of the optimization methods are conducted based on expectation. The analysis over common noise assumption, e.g. sub-Gaussian and sub-exponential is proposed by \cite{rakhlin2011making} in the strongly-convex settings, which is subsequently improved by \cite{harvey2019tight}. \cite{li2020high} prove the high probability convergence rate
for a weighted average of the squared gradient norms of SGD assuming strong smoothness and sub-Gaussian noise. \cite{madden2024high} prove a high probability bound under sub-Weibull noise, which generalizes sub-Gaussian and sub-exponential properties to heavier tailed distributions~\citep{vladimirova2020sub}. 

More recently, the community finds the heavy-tailed phenomenon are prevalent in common machine learning tasks~\citep{simsekli2019tail, reddi2020adaptive}. It is also observed in federated learning settings when the data are heterogeneous across clients~\citep{yang2022taming}. Under the heavy-tailed noise assumptions, \cite{gorbunov2020stochastic} prove the first high-probability convergence results
for Clip-SGD in the convex case, and is later generalized to Holder-continuous gradients in~\cite{gorbunov2021near}. In the case of non-convex problems, \cite{cutkosky2021high} provide a convergence bound for normalized clip-SGD. Subsequent works including~\cite{sadiev2023high} improve the bounds without bounded gradient assumptions. 

\textbf{Adaptive Learning Rates.} Adaptive learning rates are the key ingredients in deep learning optimization. Adagrad is first proposed in ~\cite{duchi2011adaptive} in aim of utilizing sparsity in stochastic gradients. Subsequent works, e.g. Adam~\citep{kingma2014adam} and AMSGrad~\citep{reddi2019convergence} have become the mainstream optimizers used in machine learning because of their superior empirical performance. These methods use implicit learning rates adaptive to the current iterate in the training process. In many cases, adaptive methods have been shown to converge faster than SGD, and with better generalization as well~\citep{reddi2019convergence}. In recent literatures, adaptive methods are shown to be capable of better dealing with the noise, which partially accounts for their empirical success.  \cite{zhang2020adaptive} show empirical connections between the noise in the gradients and Adam’s performance.  On the other hand, \cite{chezhegov2024gradient} demonstrate that AdaGrad and its delayed version can fail to converge in polynomial time under heavy-tailed noise, while adaptive clipping-based methods can cope with the noise with theoretical guarantees~\citep{zhang2020adaptive}. A combination of clipping-based methods and Adagrad is devised to achieve convergence under heavy-tailed noise in~\cite{chezhegov2024gradient}.

\section{Sketched Adaptive FL under Mild Noise}
\label{sec:mild}
In this section, we develop a generic framework for communication-efficient adaptive learning algorithms with unbiased sketching compressors, and conduct convergence analysis under bounded gradient assumptions.

\subsection{Sketched Adaptive FL (SAFL)}
A canonical federated learning setting involves $C$ clients, each associated with a local data distribution $\cD_c$. The goal is to minimize the averaged empirical risk: $\cL(x) = \frac{1}{C}\sum_{c=1}^C \mathbb{E}_{\xi \sim \cD_c} l(x, \xi)$,
where $l$ is the loss function, $x \in \mathbb{R}^d$ is the parameter vector, and $\xi$ is the data sample. We denote $\cL^c(x) = \mathbb{E}_{\xi \sim \cD_c} l(x, \xi), c \in [C]$ as the client loss computed over the local distribution. We denote $g_{t,k}^c$ as the mini-batch gradient over $\cL^c(x)$ at global step $t$ and local step $k$.


\begin{algorithm}[t]
\begin{algorithmic}
  \STATE {\bfseries Input:} Learning rate $\eta$, initial parameters $x_0$, adaptive optimizer \texttt{ADA\_OPT}
  \STATE {\bfseries Output:} Updated parameters $x_T$\\
  Initialize server moments: $m_0 = 0$, $v_0 = 0$, $\hat{v}_0 = 0$, client initial parameters: $x_{0,0}^c = x_0$, client moments: $m_{0}^c=0, v_{0}^c=0, \hat{v}_{0}^c=0, \forall c \in [C]$; \;\\
  \FOR{$t = 1, 2, \ldots, T$}
    \STATE {\bfseries Client Updates:}\\
    \FOR {$c=1,2,\ldots, C$}
        \STATE Client model synchronization: 
        \begin{small}
            $x_{t,0}^c, m_{t}^c, v_{t}^c, \hat{v}_{t}^c = \texttt{ADA\_OPT}(x_{t-1,0}^c, m_{t-1}^c, v_{t-1}^c,  \hat{v}_{t-1}^c, \barm_t)$
        \end{small}
         \\
        \FOR {$k=1,2,\ldots,K$}
        \STATE Compute stochastic gradient $g_{t,k-1}^c$ with respect to the parameters $x_{t,k-1}^c$;\\
        Perform gradient step:
        $x_{t,k}^c = x_{t,k-1}^c - \eta_t g_{t,k-1}^c$;
        \ENDFOR \\
        \STATE Sketch (compress) the parameter updates:
        \begin{small}
            \[
        \barm_{t}^c = \comp (x_{t,0}^c - x_{t,K}^c); 
        \]
        \end{small}
    \ENDFOR \\
    {\bfseries Server Updates:}\\ Average sketched client updates and send $\barm_t$ back to clients 
    \begin{small}
        \[
    \barm_t = \frac{1}{C}\sum_{c=1}^C \barm_{t}^c; 
    \]
    \end{small}
    Update paramters and moments: $x_{t}, m_{t}, v_{t}, \hat{v}_{t} = \texttt{ADA\_OPT}(x_{t-1}, m_{t-1}, v_{t-1}, \hat{v}_{t-1}, \barm_t)$.\\
  \ENDFOR \\
\end{algorithmic}
\caption{Sketched Adaptive Federated Learning (SAFL)}
\label{alg:sketch_federated}
\end{algorithm}

Algorithm~\ref{alg:sketch_federated} presents a generic framework of communication-efficient adaptive methods, which calls adaptive optimizers as subroutines. We denote $T$ as the total training rounds. At each round, after $K$ local training steps, client $c$ sends to the server the sketched local model updates with a sketching operator \sk$:\mathbb{R}^d \rightarrow \mathbb{R}^b$.
 If 
 $b \ll d$ without deteriorating the performance too much, the communication cost per round can be reduced from $O(d)$ to $O(b)$. Algorithm~\ref{alg:ams} projects the compressed updates and second moments back to the ambient dimension using a desketching operator \desk$:\mathbb{R}^b \rightarrow \mathbb{R}^d$ and implements a single-step adaptive optimization. The server and clients call Algorithm~\ref{alg:ams} at every epoch, i.e. communication round, to update the global model and synchronize local models.
 The gradient compression steps differentiate Algorithm~\ref{alg:sketch_federated} from the subspace training methods~\citep{gressmann2020improving, wortsman2021learning} since we are utilizing the global gradient vector in each round rather than solely optimizing over the manifold predefined by a limited pool of parameters. The choice of server-side optimizers determines how the lossy replicates in $\mathbb{R}^d$ are used to update the running moments (i.e. the momentum and the second moments). 
 The server sends the moments in $\mathbb{R}^b$ back to the clients so that each client can perform an identical update on its local model, which ensures synchronization as each training round starts.

\begin{remark} (Sketching Randomness).
\normalfont
At each single round, the sketching operators $\comp$'s are shared among clients, via the same random seed, 
which is essential for projecting the local model updates to a shared low dimensional subspace and making direct averaging reasonable. On the other hand, we use fresh $\comp$'s at different rounds so that the model updates lie in distinct subspaces. \qed
\end{remark}


\begin{algorithm}[t]
\begin{algorithmic}
  \STATE {\bfseries Input:}{iterate $x_{t-1}$, moments $m_{t-1}, v_{t-1}, \hat{v}_{t-1}$ , sketched updates $\bar{m}_t$}\\
  \STATE {\bfseries Parameters:}{Learning rate $\lr$, $\beta_1$, $\beta_2$, Small constant $\epsilon$}\\
  \STATE {\bfseries Output:}Updated parameters $x_{t}$, and moments $m_t$, $v_t$, $\hat{v}_t$ \\
  
    Update ~ $m_t = \beta_1 \cdot m_{t-1} + (1 - \beta_1) \cdot \decomp(\bar{m}_t)$;
    
    Update ~ $v_t = \beta_2 \cdot v_{t-1} + (1 - \beta_2) \cdot \decomp(\bar{m}_t)^2$;
    
    Update ~ $\hat{v}_t = \max(\hat{v}_{t-1}, v_t).$
    
    Update ~
    $x_{t+1} = x_{t} - \frac{\lr}{\sqrt{\hat{v}_t} + \epsilon} \cdot m_t := x_t - \kappa {\hat{V}_{t}}^{-1/2} m_t$.
\end{algorithmic}
\caption{\texttt{ADA\_OPT} (AMSGrad)}
\label{alg:ams}  
\end{algorithm}

\subsection{Random Sketching}
We will first introduce the desired characteristics of compression and then list a family of sketching algorithms which possess those properties.

\begin{property} (Linearity). The compression operators are linear w.r.t the input vectors, i.e. $\comp(\sum_{i=1}^n {v_i}) = \sum_{i=1}^n {\comp(v_i)}$  and   $\decomp(\sum_{i=1}^n {\bar{v}_i}) = \sum_{i=1}^n {\decomp(\bar{v}_i)},~~ \forall \{v_i, \bar{v}_i \in \mathbb{R}^d \}_{i=1}^n$.
\end{property}

\begin{property} (Unbiased Estimation).
    For any vector $v \in \mathbb{R}^d$, $\E[\decomp(\comp(v))] = v$.
\end{property}

\begin{property} (Bounded Vector Products).
\label{property_vector_product}
    For any fixed vector $v, h \in \mathbb{R}^d$, $\P(\vert \langle \decomp(\comp(v)), h \rangle - \langle v, h \rangle \vert \ge (\logd)\norm{v}\norm{h}) \le \Theta(\delta)$.
\end{property}

Property 1 and 2 guarantee the average of first moments in Algorithm~\ref{alg:sketch_federated} over clients are, in expectation, the same as those in \texttt{FedOPT}. Property 3 quantifies the bound on the deviation of vector products when applying compression. $\comp(v) = Rv$ and $\decomp(\bar{v}) = R^\top \bar{v}$, where $R \in \R^{b \times d}$ is a random sketching operator, satisfy all the properties above~\citep{song2023sketching}. We denote $R_t$ as the sketching operator used in round $t$. 
Different instantiations of $R$ constitute a rich family of sketching operators, including i.i.d.~isotropic Gaussian~\citep{song2023sketching}, Subsampled Randomized Hadamard Transform (SRHT)~\citep{lu2013faster}, and Count-Sketch~\citep{charikar2002finding}, among others. The specific error bounds for these special cases can be found in Appendices~\ref{lemma_srht}, \ref{lemma_gaussian}, and \ref{lemma_count_sketch} respectively.


    

\subsection{Convergence Analysis}
We first state a set of standard assumptions commonly used in the literature of first-order stochastic methods. We will use $\Vert \cdot \Vert$ to denote $L_2$-norm throughout the work. 

\begin{assumption} (Bounded Global Gradients). Square norm of the gradient is uniformly bounded, i.e., $\Vert \nabla \cL(x) \Vert^2 \le G_g^2$.
\label{a:bgg}
\end{assumption}


\begin{assumption} (Bounded Client Gradients).
For every client, there exists a constant $G_c \ge 0$,  such that $\Vert \nabla \cL^c(x) \Vert^2 \le G_c^2, ~c \in [C]$.
\label{a:bcg}
\end{assumption}

For simplicity, in this section we define $G := \max \{ \max\{G_c\}_{c=1}^C, G_g\}$ to denote the upper bound for client and global gradient norms. We further show in Section~\ref{sec:nogradnorm} that Assumption $\ref{a:bgg}$ and \ref{a:bcg} can be removed when deriving convergence bound. We assume the local stochastic noise from mini-batches is sub-Gaussian, which is widely adopted in first-order optimization~\citep{harvey2019tight,mou2020linear}. 

\begin{assumption} (Sub-Gaussian Noise).
The stochastic noise $\norm{\nabla \cL^c(x) - g^c(x)}$ at each client is a $\sigma$-sub-Gaussian random variable, i.e. $\P(\norm{\nabla \cL^c(x) - g^c(x)} \ge t) \le 2 \exp(-t^2/\sigma^2)$, for all $t\ge 0$.
\label{assume_local_grad_sub}
\end{assumption}

Besides, we have assumptions on the Hessian eigenspectrum $\{\lambda_i, v_i\}_{i=1}^d$ of the loss function $\cL$.
\begin{assumption} (Hessian Matrix Eigenspectrum)
    The smoothness of the client loss function $\cL_i$, i.e. the largest eigenvalue of the loss Hessian $H_{\cL_i}$ is bounded by $L$.
    \label{assume_hessian}
\end{assumption}
The local smoothness assumption is commonly used in federated learning settings~\citep{safaryan2021fednl, fatkhullin2024momentum} and holds for general deep learning losses. It can be directly derived from Assumption~\ref{assume_hessian} that the global loss $\cL = \frac{1}{C}\sum_{c=1}^C \cL_c$ is $L-$smooth. 

\begin{definition} (Intrinsic Dimension)
    Let $\{\lambda_i\}_{i=1}^d$ be the eigenspectrum of the loss Hessian $H_\cL$. The intrinsic dimension is defined as  $\intdim = \sum_{i=1}^d \vert \lambda_i \vert / \max_i \vert \lambda_i \vert$.
\end{definition}
The definition of intrinsic dimension is analogous to what is proposed in~\cite{ipsen2024stable}, where we take the absolute values of eigenvalues. Intuitively, the Hessian matrix possesses an anisotropic structure in different directions, whereas the convectional smoothness is a pessimistic estimation of the loss curvature. A large volume of recent literature has indicated that the intrinsic dimension of the Hessian in deep learning models  can be significantly smaller than the ambient dimensionality $d$. \citep{ghorbani2019investigation, li2020hessian, liu2023sophia} show the eigenspectrum enjoys a sharp decay in magnitude. \citep{sagun2016eigenvalues, liao2021hessian} show the eigenspectrum have bulk parts concentrate at zero. ~\citep{xie2022power,zhang2024transformers} further show the eigenvalues conform with a power-law distribution, and in this case the intrinsic dimension is a constant independent of $d$.
We quote their plots in Appendix~\ref{app:expts} for completeness. Our empirical verification under the setting of FL can also be found in Fig.~\ref{fig:vit-small-eigenspectrum} in Appendix~\ref{app:expts}.

\begin{remark}
(Three types of noises in Algorithm~\ref{alg:sketch_federated}).
\normalfont
One of the key technical contributions of this work is to theoretically balance the noises of different sources and derive a reasonable convergence rate which is independent of the number of parameters. The noise in the training process stems from the local mini-batch training, the compression error due to sketching, and the aggregate noise over the training horizon. The stochastic error of mini-batch training is $\sigma$-sub-Gaussian by Assumption~\ref{assume_local_grad_sub}. We will adopt a probability variable $\delta_g$, which is usually viewed as a tiny value (1e-5), to yield a high probability bound on the sub-Gaussian noise. 
The sketching error depends on the specific choice of sketching methods, but is always controlled by the bounded property on vector products (Property~\ref{property_vector_product}). Analogous to $\delta_g$, we denote the probability variable in sketching as $\delta$.  The two kinds of noise are unbiased and additive to the gradient, and have sequential dependencies. In the analysis (Appendix~\ref{app:mild}), we will introduce a martingale defined over the aggregated noise, using which we can derive a high-probability concentration bound for the variance. We denote $\martin$ as the scale of the $\psi_2$-norm~\citep{vershynin2018high} in the martingale. \qed 
\end{remark}

Now we characterize the convergence of Algorithm~\ref{alg:sketch_federated} in Theorem~\ref{theorem_mild}. All technical proofs for this section are in Appendix~\ref{app:mild} and we provide an outline of the proof techniques in Section~\ref{ssec:psketch}.
\begin{theorem}
    Suppose the sequence of iterates $\{x_t\}_{t=1}^T$ is generated by Algorithm~\ref{alg:sketch_federated} (SAFL) with a constant learning rate $\eta_t \equiv \eta$. Under Assumptions 1-4, for any $T$ and $\epsilon > 0$, with probability $1 - \Theta(\delta) - O(\exp(-\Omega(\martin^2))) -\delta_g $,
    \begin{small}
    \begin{align*}
     \lr \eta J_1 K \sum_{t=1}^T \Vert \nabla \cL (x_t) \Vert^2  
    \leq & \cL(z_1) + \frac{1}{\epsilon} \lr \eta^2 L K^2G^2 T + \martin \lr \eta K\sqrt{T}(\logtd \frac{G^2}{\epsilon} + \frac{\sigma}{\epsilon}\log^{\frac{1}{2}}(\frac{2T}{\delta_g})) 
    \\
    &+ \eta^2 \lr T (1+\logtsqd)^2 \frac{8\lr \intdim L K^2 + 2G}{(1-\beta_1)^2} \frac{G^2}{\epsilon^2},
    \end{align*}   
    \end{small}
    where $\delta, \delta_g$, and $\martin$ are the randomness of sketching, sub-Gaussian noise, and martingales respectively, and $J_1 := \left(\sqrt{1+\logtdsq} \eta K G + \epsilon \right)^{-1}$.
    \label{theorem_mild}
\end{theorem}

\begin{remark} (Dependence on $K$)
    The bound in Theorem~\ref{theorem_mild} has a dependence on $K$. The primary focus of this work is to reduce the communication cost in FL algorithms, where the cost only depends on $T$ and compression rate ($b)$. Therefore, we view $K$ as a constant throughout the work. As we will show in Corollary \ref{corollary_mild} and \ref{corollary_mild_near_init},  if we set $\eta$ as $O(1/K\sqrt{T})$, which is the same as \citet{reddi2020adaptive}, the dependence on $K$ in the bound can be eliminated.
\end{remark}

A non-asymptotic convergence bound of training with practical decaying learning rates can be found in Theorem~\ref{theorem_mild_non_asymptotic} in appendix. Given that we only introduce logarithmic factors on $d$ in the iteration complexity and the per-round communication $b$ is a constant, the total communication bits of training a deep model till convergence is also logarithmic w.r.t $d$. To better understand Theorem~\ref{theorem_mild}, we can investigate different regimes based on the training stages. For the asymptotic regime, where $T$ is sufficiently large, we can achieve an $O(1/\sqrt{T})$ convergence rate in Corollary~\ref{corollary_mild}.
\begin{corollary}
     (Asymptotic Regime of Theorem~\ref{theorem_mild}) With the same condition as in Thereom~\ref{theorem_mild} and a constant learning rate $\eta_t \equiv \frac{1}{K\sqrt{T}}$, 
     for  sufficiently large $T \ge \frac{G^2}{\epsilon^2}$, with probability $1 - \Theta(\delta) - O(\exp(-\Omega(\martin^2))) -\delta_g $,
    \begin{small}
    \begin{align*}
    \frac{1}{T}\sum_{t=1}^T& \Vert \nabla \cL (x_t) \Vert^2  
    \leq   \frac{4}{\sqrt{T}} (1+J_2)^2 \frac{4\lr \intdim L + G}{(1-\beta_1)^2} \frac{ G^2}{\epsilon}+\frac{2\cL(z_1)\epsilon}{\kappa\sqrt{T}} + \frac{2}{\epsilon}  \frac{L G^2}{\sqrt{T}}  + \martin \frac{2}{\sqrt{T}}(J_2 G^2 + \sigma\log^{\frac{1}{2}}(\frac{2T}{\delta_g})) ,
    \end{align*}   
    \end{small}
    where $\delta, \delta_g$ and $\martin$,  are the randomness of sketching, sub-Gaussian noise and martingales respectively, and $J_2:=\logtsqd$
    \label{corollary_mild}
\vspace*{-3mm}
\end{corollary}
More interestingly, in the near-initialization regime, where $T$ is relatively small, we can observe that the coefficient of $\Vert \nabla \cL (x_t)\Vert^2$ on the left hand side in Theorem~\ref{theorem_mild} and \ref{theorem_mild_non_asymptotic} is approximately a constant, given that $\epsilon$ is tiny. Therefore, SAFL can achieve an $O(1/T)$ convergence near initialization, which accounts for the faster convergence speed than non-adaptive methods. 

\begin{corollary}
     (Near-initialization Regime of Theorem~\ref{theorem_mild}) With the same condition as in Thereom~\ref{theorem_mild} and a constant learning rate $\eta_t \equiv \frac{1}{K\sqrt{T}}$, set $b\ge \log^3(CKd^2T^2/\delta)$ and constant $J_3 > \sqrt{2}G$, then for any $T \le \frac{J_3-\sqrt{2}G}{\epsilon^2}$, with probability $1 - \Theta(\delta) - O(\exp(-\Omega(\martin^2))) -\delta_g $,
    \begin{small}
    \begin{align*}
    \frac{1}{J_3T}\sum_{t=1}^T &\Vert \nabla \cL (x_t) \Vert^2  
    \leq \frac{\cL(z_1)\epsilon}{\kappa T} + \frac{1}{\epsilon}  \frac{L G^2}{T} + \frac{\martin }{T}(G^2 + \sigma\log^{\frac{1}{2}}(\frac{2T}{\delta_g})) +  \frac{8}{T}\frac{4\lr \intdim L + G}{(1-\beta_1)^2} \frac{G^2}{\epsilon},
    \end{align*}   
    \end{small}
    where $\delta, \delta_g$ and $\martin$ are the randomness of sketching, sub-Gaussian noise and martingales respectively.
    \label{corollary_mild_near_init}
\vspace*{-3mm}
\end{corollary}

\subsection{Technical Results and Proof Sketch}
\label{ssec:psketch}
In this section, we provide a sketch of the proof techniques behind the main results. We focus on the proof of Theorem~\ref{theorem_mild}, and the proof of Theorem~\ref{theorem_mild_non_asymptotic} shares the main structure. The proof of Theorem~\ref{theorem_mild} contains several critical components, which are unique to adaptive methods. We follow the common proof framework of adaptive optimization, and carefully deal with the noise introduced by random sketching in the momentum. We adopt AMSGrad (Alg.~\ref{alg:ams}) as the server optimizer and it would be straightforward to extend the analysis to other adaptive methods.

We first introduce the descent lemma for AMSGrad. For conciseness, we denote the precondtioner matrix ${\rm diag}((\sqrt{\hat{v}_t}+\epsilon)^2)$ as $\hat{V}_t$. Define an auxiliary variable $z_t = x_t + \frac{\beta_1}{1-\beta_1}(x_t - x_{t-1})$. The trajectory of $\cL$ over $\{z_t\}_{t=1}^T$ can be tracked by the following lemma.
\begin{lemma}
\label{lemma_descent_amsgrad_informal} (Informal version of Lemma~\ref{lemma_descent_amsgrad})
    For any step $t \in [T]$, 
    \begin{small}
    \begin{align*}
    \cL(z_{t+1}) \lessapprox &\cL(z_t) - \frac{\kappa\etat}{C}\sum_{c=1}^C  \sum_{k=1}^K\nabla \cL(x_t)^\top {\hat{V}_{t-1}}^{-1/2}   R_t^\top R_t g_{t,k}^c +  (z_t - x_t)^\top H_{\cL}(\hat{z}_t)(z_{t+1} - z_t),
    \end{align*}
    \end{small}
    where $ H_{\cL}(\hat{z}_t)$ is the loss Hessian at some $\hat{z}_t$ within the element-wise interval of $[x_t, z_t]$, and $\lessapprox$ omits the less important terms.
\end{lemma}
Our objective henceforth is to bound the first-order descent term and the second-order quadratic term on the right hand side respectively.

\textbf{Second-Order Quadratic Term.} Denote $\{\lambda_j, v_j \}_{j=1}^d$ as the eigen-pairs of $ H_{\cL}(\hat{z}_t)$. The quadratic term can be written as $(z_t-x_t)^\top  H_{\cL}(\hat{z}_t) (z_{t+1}-z_t) = \sum_{j=1}^d \lambda_j \langle z_{t+1}-z_t, v_j \rangle \langle z_{t}-x_t, v_j \rangle.$
The inner product terms can be viewed as a projection of the updates onto anisotropic bases. Since $z_{t+1} - z_t$ and $z_t-x_t$ can both be expressed by $x_{t+1}-x_t$ and $x_t - x_{t-1}$, we can bound the quadratic term using the following lemma.
\begin{lemma} For any $t \in [T]$, $\vert \langle x_{t}-x_{t-1}, v_j \rangle \vert \le \kappa \eta (1+\frac{\log^{1.5}(CKtd/\delta)}{\sqrt{b}})\frac{K G}{\epsilon}$, with probability $1-\delta$.
    \label{lemma_inner_product}
\vspace*{-3mm}
\end{lemma}
Bounding the inner-product term is non-trivial since $z_t$ contains momentum information which depends on the randomness of previous iterations. A proof of a generalized version of this statement is deferred to the appendix, where induction methods are used to address the dependence. Combining Lemma~\ref{lemma_inner_product} with Assumption~\ref{assume_hessian} yields a dimension-free bound on the second-order quadratic term.

\begin{remark}
\normalfont
A straightforward application of smoothness to the second-order term yields a quadratic term $\norm{R^\top Rg}^2$, which is linearly proportional to $d$ in scale \citep{rothchild2020fetchsgd,song2023sketching}. We avoid this dimension dependence by combining Property~\ref{property_vector_product} of sketching and the intrinsic dimension of deep learning Hessian.
\qed 
\end{remark}

\textbf{First-Order Descent Term}. 
The first-order term in the descent lemma can be decomposed into three components, which we will handle separately:
\begin{small}
\begin{align*}    \nabla \cL(x_t)^\top {\hat{V}_{t-1}}^{-1/2}   R_t^\top R_t g_{t,k}^c = &\underbrace{\nabla \cL(x_t)^\top {\hat{V}_{t-1}}^{-1/2} \nabla \cL^c(x_t)}_{\cD_1^c} +   \underbrace{\nabla \cL(x_t)^\top {\hat{V}_{t-1}}^{-1/2} (R_t^\top R_t g_{t,k}^c - \nabla \cL^c(x_{t,k}^c))}_{\cD_2^c} \\
    &+  \underbrace{\nabla \cL(x_t)^\top {\hat{V}_{t-1}}^{-1/2}(\nabla \cL^c(x_{t,k}^c) - \nabla \cL^c(x_t))}_{\cD_3^c}.
    \label{eq:fod}
\end{align*}
\end{small}
First, $\cD_3^c$ 
can be reduced to a second-order term by smoothness over $\cL$, $\nabla \cL(x_t)^\top {\hat{V}_{t-1}}^{-1/2}(\nabla \cL^c(x_{t,k}^c) - \nabla \cL^c(x_t))
    =  -\eta \sum_{\tau=1}^k \nabla \cL(x_t)^\top {\hat{V}_{t-1}}^{-1/2}\hH_\cL^c g_{t, \tau}^c.$
%
Note that this term does not involve any stochasticity from random sketching, hence we can directly derive the upper bound by Cauchy-Schwartz. Next, since $\sumv{C} \nabla \cL^c(x_t) = \nabla \cL(x_t)$, $\cD_1^c$ composes a scaled squared gradient norm. Applying element-wise high probability bound on random sketching yields the lower bound for the scale.
\begin{lemma} For ${\hat{V}_{t-1}}^{-1/2}$ generated by Algorithm~\ref{alg:sketch_federated} 
 (SAFL), with probability $1-\delta$,
\begin{align*}
\nabla \cL(x_t)^\top {\hat{V}_{t-1}}^{-1/2} \nabla \cL(x_t) 
\ge M^{-1} \Vert \nabla \cL (x_t) \Vert^2, 
\end{align*}
where $M = \sqrt{1+\frac{\log^{1.5}(CKtd^2)}{\sqrt{b}}} \eta K G + \epsilon$.
\label{lemma_descent_bound}
\end{lemma}
%
%
\textbf{Martingale for zero-centered noise.} $\cD_2^c$ contains a zero-centered noise term $R_t^\top R_t g_{t,k}^c - \nabla \cL^c(x_{t,k}^c)$, where the randomness is over $R_t$ and the mini-batch noise at round $t$. Although $x_{t,k}^c$ has temporal dependence, the fresh noise due to mini-batching and sketching-desketching at round $t$ is independent of the randomness in the previous iterations. Therefore, the random process defined by the aggregation of the zero-centered noise terms over time forms a martingale. The martingale difference can be bounded with high probability under our proposed sketching method. Then by adapting Azuma's inequality on a sub-Gaussian martingale, we have
\begin{lemma}
\label{lemma_martingale}
    With probability $1- O(\exp(-\Omega(\martin^2))) -\delta -\delta_g$,
\begin{small}
\begin{align*}
    &\sum_{t=1}^T  \left\vert \frac{1}{C}\sum_{c=1}^C  \sum_{k=1}^K \nabla \cL(x_t)^\top {\hat{V}_{t-1}}^{-1/2} (R_t^\top R_t g_{t,k}^c - \nabla \cL^c(x_{t,k}^c)) \right\vert     \le \martin \sqrt{T}(\logtd \frac{KG^2}{\epsilon} + \frac{\sigma}{\epsilon}\log^{\frac{1}{2}}(\frac{2T}{\delta_g})).
\end{align*}
\end{small}
\end{lemma}
Finally, applying union bounds to these parts and telescoping  the descent lemma leads to Theorem~\ref{theorem_mild}.

\label{headings}


\section{Bounded Gradient Norm Along Optimization Trajectory}
\label{sec:nogradnorm}
Although the gradient norm assumptions (Assumption~\ref{a:bgg} and \ref{a:bcg}) are standard and mild assumptions in adaptive optimization~\citep{reddi2020adaptive} and federated learning research~\citep {basu2019qsparse, xie2020cser}, these assumptions might not hold in the ambient space for neural network loss. In this section, we show that the two assumptions are not necessary to derive the convergence bound.

\begin{figure*}[t]
\vspace*{-3mm}
    \centering
    \includegraphics[scale=0.4]{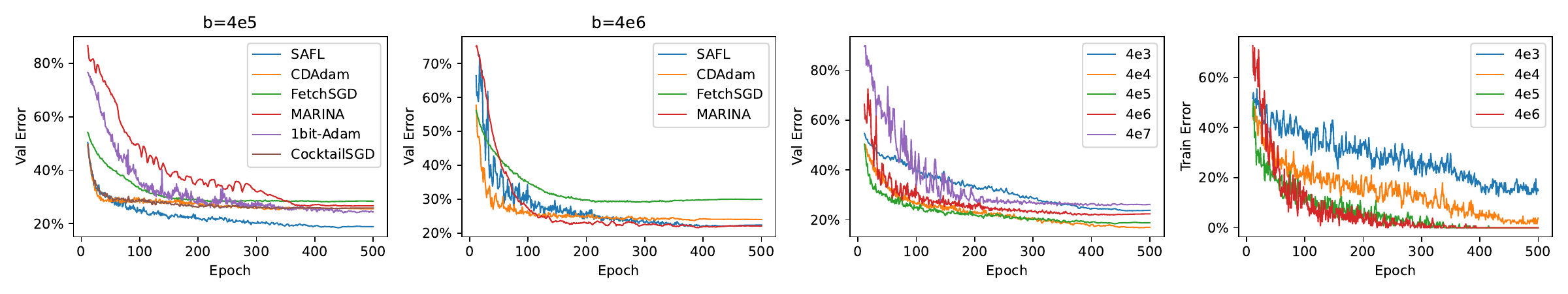}
    \vspace*{-3mm}
    \caption{Model performance on CIFAR-10 with ResNet of 42M parameters. The plot starts from the 10th epoch for better demonstration; Third: Validation error on SAFL with different sketch sizes. The legend 4e7 represents training in the ambient dimension without sketching. Fourth: Training error on SAFL with different sketch sizes. Larger sketch size improves the convergence rate and the peak validation error is achieved when $b=4e4$.}
    \label{fig:cifar-sketch}
\end{figure*}

Our approach is to show the gradient norm is bounded over the entire optimization path with high probability. We rely on the following lemma  to demonstrate the boundedness.
\begin{lemma}
    \label{lemma_grad_bound_smooth}
    For any $L$-smooth function $\cL(x)$ with optimal value $\cL^* \ge 0$, $\Vert \nabla \cL(x)\Vert^2 \le 2L\cL(x)$.
\end{lemma}
As stated in Lemma~\ref{lemma_grad_bound_smooth}, for any smooth function, the gradient norm can be bounded by the function value at the specific iterate. That being said, we can derive an upper bound on the gradient norm along the optimization trajectory via bounding the function values over the iterates. However, the technical difficulty of the analysis lies in the involvement of the local training steps, which might be noisy and the relation of which with the global iterate is unclear. 

Our analysis can be divided into two steps: 1) We first relate the averaged local gradient norm to the global function value based on the local smoothness. Notice that this step does not require any additional assumptions, such as the deviation between local and global function values; 2) We apply the induction method to show the global loss is contained in the neighborhood of the function value at initialization, and the bound of the gradient norm follows immediately by applying Lemma~\ref{lemma_grad_bound_smooth}. 

The following lemma shows how the local gradient norm can be related to the local loss at the global iterate $x_t$,
\begin{lemma}
    Under Assumption~\ref{assume_local_grad_sub},  Let $\etat \le \frac{1}{2L \sqrt{K}}$. The local gradients as of $k \le K$ can be bounded by 
    \begin{small}
        \begin{align*}
        \Vert \nabla \cL^c (x_{t,k}^c)\Vert \le \sqrt{2\localnoise^2 \ln \frac{2}{\stocdelta}} + \sqrt{2\localnoise^2 \ln \frac{2}{\stocdelta}+4L \cL^c(x_t)+\localnoise^2},
        \end{align*}
    \end{small}
    with probability $1-K \stocdelta - K \exp(-\localnoise^2/\sigma^2)$.
\end{lemma}
Applying the fact that $\cL(x^t) = \frac{1}{C}\sum_{c=1}^C \cL^c(x_t)$, the averaged local gradient can be bounded by the global loss,
\begin{align*}
    \frac{1}{C} \sum_{c=1}^C \Vert  \nabla \cL^c(x_{t,\tau}^c)\Vert \le 2\sqrt{L} \sqrt{\cL(x_t)} + 2\sqrt{2\localnoise^2 \ln \frac{2}{\stocdelta}} + \localnoise.
\end{align*}
The averaged local gradient norm will be a key component in the following analysis that focuses on the global gradients.

Suppose the induction basis is $ \cL(x_\tau) \le \frac{G}{2L}$ for $\tau \le t$ with high probability, for some $G$ that will be specified later. 
We revisit the terms in Section~\ref{sec:mild}. For instance, when the condition in the induction basis holds,
\begin{small}
   \begin{align*}
\frac{1}{C}\sum_{c=1}^C\cD_3^c  \le &  \frac{\lr\etat^2 L}{C} \Vert \nabla \cL(x_t)\Vert \Vert \hat{V}_{t-1}^{-1/2} \Vert \sum_{c=1}^C  \sum_{k=1}^K  \Vert \sum_{\tau=1}^{k-1} g_{t,\tau}^c\Vert \\
    \le & \frac{\sqrt{2}\lr\etat^2 K^2 L^2}{\epsilon} \cL(x_t) + \frac{2\lr\etat^2 K^2 L}{\epsilon} \Vert \nabla \cL(x_t)\Vert \localnoise (1+ \sqrt{2\ln \frac{2}{\stocdelta}}),
\end{align*} 
\end{small}
with probability $1-CK\stocdelta - CK \exp(-\localnoise^2/\sigma^2)$. $\cD_2^c$ can be dealt with in a similar way by constructing a martingale. The key observation in the above bound is that $\frac{1}{C}\sum_{c=1}^C\cD_3^c$ is quadratic in $\eta$. With the specific choice of $\eta=\frac{\eta_0}{\sqrt{T}}$ where $\eta_0$ is a constant, the term over time step $T$ is summable, i.e. $\sum_{t=1}^T \frac{1}{C}\sum_{c=1}^C\cD_3^c$ is a constant. Likewise, we can show that the other terms are summable and leads to an upper bound of $\cL(z_T)$ by the following lemma.

\begin{lemma}
    We can derive an upper bound on $\cL(z_{T})$
    \begin{small}
    \begin{align*}
        &\cL(z_{T}) \le \lr \etat \sqrt{T} \cM_1 G + \lr \etat \sqrt{T} \cM_2 \sqrt{G} + \cM_3  +\sum_{t=1}^{T-1} (\lr \etat^2 \cM_4 G^{3/2} + \lr \etat^2 \cM_5 G + \lr \etat^2 \cM_6\sqrt{G}+\lr^2\etat^2 \cM_7 G),      
    \end{align*}
\end{small}
where $\{\cM_i\}_{i=1}^7$ are constants independent of $\lr, \etat$ and $G$, and the full forms can be found in Appendix~\ref{app:nogradnorm}.
\end{lemma}
With the closeness of $z_T$ and $x_T$, we can show $\cL(x_T) \le \frac{G}{2L}$ under appropriate choice of $\lr$ and $\eta$, and it is sufficient to ensure both the induction basis holds and the gradient norm is bounded.  The upper bound of $\Vert \nabla \cL(x_t) \Vert^2$ over the entire optimization path is provided by the following theorem.
\begin{theorem}
    Let $G := \max \{ 2 \localnoise^2 (1+ \sqrt{2\ln \frac{2CK}{\stocdelta}})^2,\tilde O(1)/\sqrt{b} + \cM \}$ where the full form can be found in \eqref{eq:g_bound} in  Appendix~\ref{app:nogradnorm}, $\lr = \frac{1}{\sqrt{G}}$, and $\etat = \frac{\eta_0}{\sqrt{T}}$ subject to
     $\eta_0 \le \min\{\frac{\epsilon}{6\sqrt{L}}  (1+\frac{\log^{1.5}(CKTd^2/\delta)}{\sqrt{b}})^{-1}, \frac{\sqrt{T}}{2L\sqrt{K}} \}.$
    Then under Assumption~\ref{assume_local_grad_sub} and \ref{assume_hessian}, with probability $ 1- T{\rm exp}(-\Omega(\martin^2))- T C\stocdelta - TCK\exp(-\localnoise^2/\sigma^2) - T\delta$, the gradient on the iterates $x_t$ generated by Algorithm~\ref{alg:sketch_federated} are bounded by $G$, i.e. $\Vert \nabla \cL(x_t) \Vert^2 \le G, ~~ t\le T.$
    Consequently, the averaged gradient converges with rate $O(1/\sqrt{T})$ by
    \begin{small}
        \begin{align*}
        \frac{1}{T} \sum_{t=1}^T \Vert \nabla \cL(x_t) \Vert^2 \le \frac{G}{2\kappa \eta_0 L^2\sqrt{T}}\left(\eta K \cM_8  + \epsilon \right),
    \end{align*}
    \end{small}
    where $\cM_8 := \sqrt{1+\logtdsq} (\sqrt{2G}+2 \localnoise(1+ \sqrt{2\ln \frac{2}{\stocdelta}}))$.
\label{theorem_no_grad_norm}
\end{theorem}

\section{Empirical Studies}
\label{sec:expts}
In this section, we instantiate the algorithm framework of SAFL to demonstrate the effect of sketching in common federated deep learning settings.

\begin{figure*}[t]
\vspace*{-3mm}
    \centering
    \includegraphics[scale=0.50]{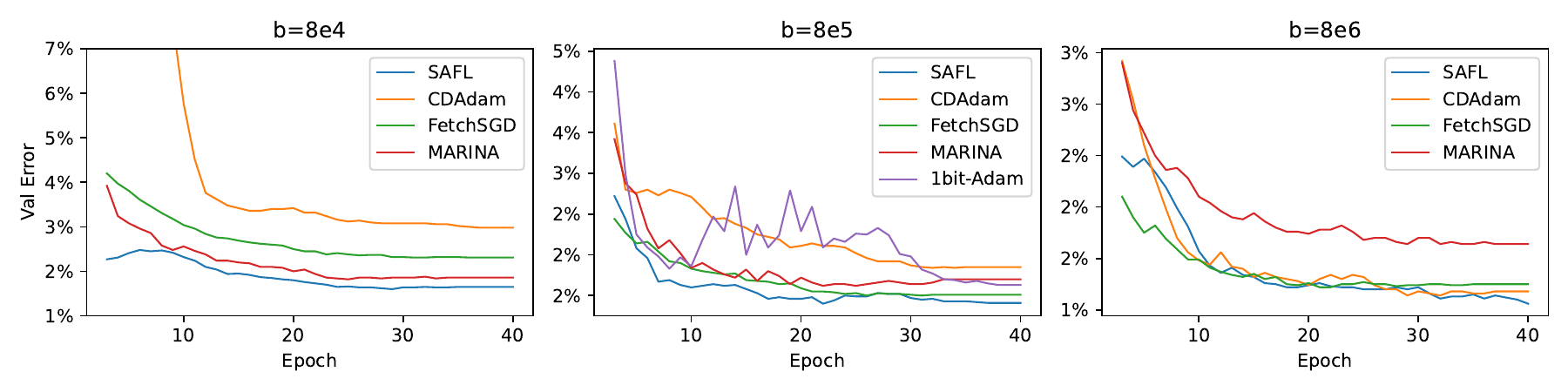}
    \vspace*{-3mm}
    \caption{Validation Error on CIFAR-10. We finetune a ViT-base model (with 86M parameters) from the pretrained backbone checkpoint~\citep{dosovitskiy2020image}.  1Bit-Adam has comparable compression rates with $b=8e5$. SAFL optimizer consistently outperforms in all sketch sizes.}
    \label{fig:cifar-vit-sketch}
\end{figure*}

\begin{figure*}[t]
\vspace*{-3mm}
    \centering
    \includegraphics[scale=0.50]{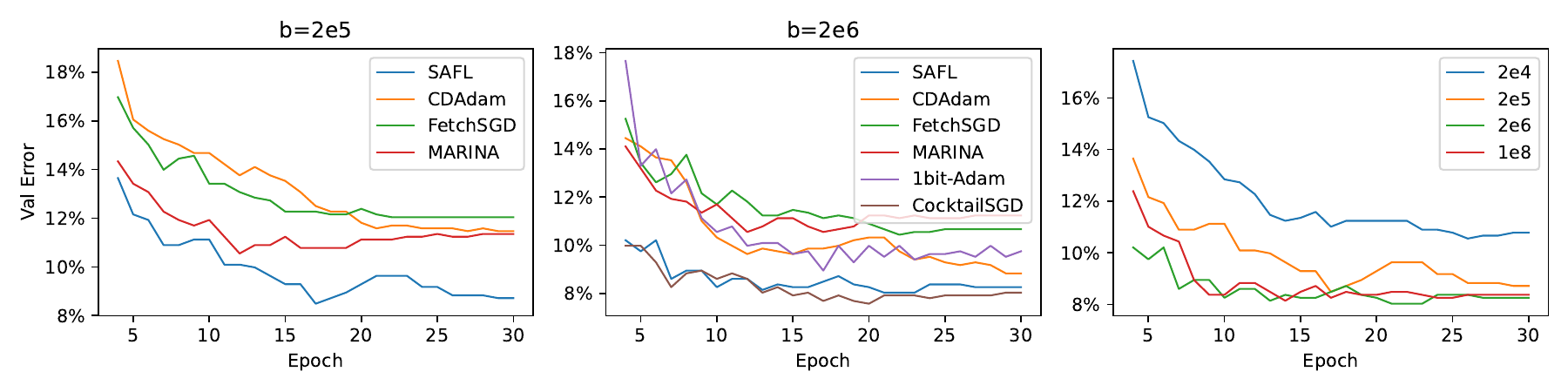}
    \vspace*{-3mm}
    \caption{Validation Error on SST2 (GLUE) with BERT of 100M parameters. Left: sketch size $b=2e5$; Middle: $b=2e6$; Right: \texttt{ADA\_OPT} is Adam, with sketch size $b \in \{2e4, 2e5, 2e6\}$. The legend $1e8$ represents training in the ambient dimension without sketching. Larger sketch sizes mainly improves the convergence rate and achieve comparable test errors at the end of training.} 
    \label{fig:sst-sketch}
\end{figure*}

\textbf{Experimental Configurations.}
We adopt three distinct experimental settings, from vision to language tasks, and in finetuning and training-from-scratch regimes. For the vision task, we train a ResNet101~\citep{wu2018group} with a total of 42M parameters from scratch and finetune a ViT-Base~\citep{dosovitskiy2020image} with 86M parameters on the CIFAR-10 dataset~\citep{krizhevsky2009learning}. 
For the language task, we adopt SST2, a text classification task, from the GLUE benchmark~\citep{wang2018glue}. We train a BERT model~\citep{devlin2018bert} which has 100M parameters.
For all experiments we split the training dataset uniformly over 5 clients. We adopt Adam as the base adaptive optimizer at the server side. 
We select representative approaches as baselines methods, including  FetchSGD~\citep{rothchild2020fetchsgd}, MARINA~\citep{gorbunov2021marina}, CocktailSGD~\citep{wang2023cocktailsgd}, CDAdam~\citep{wang2019adaptive} and 1 bit-Adam~\citep{tang20211}. A comparison on the iteration complexity and communication costs of the baseline methods can be found in Table~\ref{tab:theory_compare} in the appendix. We use the term sketch size $b$ to denote the uplink communication bits in each round. Some methods, such as MARINA, may take higher downlink communication cost. We reaffirm that our main target is to show SAFL is competitive in performance with better theoretical guarantees, but not to beat the existing algorithms well-suited for production.


\textbf{Sharp-Decaying Hessian Eigenspectrum}. Our theoretical result builds upon the notion of intrinsic dimension. While existing research has repeatedly shown supporting evidence on the sharp-decaying eigenspectrum, we also provide an affirmative verification in the context of federated deep learning in Fig.~\ref{fig:vit-small-eigenspectrum} in the Appendix.



\textbf{Sketched Adaptive FL.} 
Fig.~\ref{fig:cifar-sketch} depicts the error curve on the validation set of CIFAR-10 when training ResNet(40M) with sketch sizes $b \in \{4e5, 4e6\}$. The compression rate of 1bit-Adam is fixed at 97\%, which is comparable with the compression rate 99\% achieved at $b=4e5$. CocktailSGD also achieves a 99\% compression rate under its default parameters. We plot the curve of validation error to get a better sense of the convergence speed of each algorithm. We can see for sketch size=4e5, our SAFL outperforms other optimizers in the validation error by a significant margin. For sketch size=4e6, SAFL and MARINA performs alike and outperform FetchSGD and CDAdam. More interestingly, we compare the model performance of SAFL with different sketch sizes and find that in this experimental setting, the validation error is not monotonic with the sketch size and reaches the peak value when $b=4e4$. On the other hand, the training error, which better reflects the convergence speed, is strictly monotonic with sketch sizes -- larger sketch size leads to faster convergence and agrees with our theory. The discrepancy between the two rates indicates sketching methods may be implicitly improving the model generalization ability.


Similar phenomenon is observed in the language task. Fig.~\ref{fig:sst-sketch} shows the test errors of training SST2 with BERT (100M parameters). The sketch sizes are selected from $\{2e5, 2e6\}$. We observe SAFL converges faster and achieves a slightly better validation performance at sketch size 2e5. At sketch size 2e6, the model performance is comparable with cocktailSGD and consistently outperforms other algorithms. We also compare SAFL with different sketch sizes from $\{2e4, 2e5, 2e6\}$, and observe the SAFL algorithm generally converges faster with larger sketch sizes. Note that the sketch size of $2e4$ (20K) is tiny, given that the ambient dimension is 100M. It is quite thrilling that using an extremely high compression rate (99.98\%), the model can still achieve comparable performance as trained in the ambient dimension. We also present results on finetuning a ViT-Base model (80M parameters) in Fig.~\ref{fig:cifar-vit-sketch}.  The sketch size $b \in \{8e4, 8e5, 8e6\}$. We see, in the finetuning regime, the SAFL optimizer achieves better performance compared with all baseline methods. 

We additionally experiment with extremely low compression rates to show the logarithmic dependence can be empirically grounded. The experiments are conducted under the same setting as Figure~\ref{fig:cifar-sketch} and \ref{fig:sst-sketch} respectively. We adopt sketch size $b \in \{4\times 10^2, 4\times 10^3, 4\times 10^4, 4\times 10^5\}$ in the CIFAR-10 task and $b\in \{2 \times 10^3, 2 \times 10^4, 2 \times 10^5, 2 \times 10^6\}$ in the SST-2 task. We present the validation errors along the training process in Figure~\ref{fig:sketch_tiny} in Appendix~\ref{app:expts}. We observe that although the validation accuracies converge to distinct values, the convergence holds for all sketch sizes. More interestingly, the convergence speed for different sketch sizes are comparable. Even under extremely tiny sketch sizes, SAFL converges in the first 100 (25 \textit{resp.}) epochs in CIFAR-10 (SST2 \textit{resp.}) task. This observation aligns with our theoretical results on the logarithmic dependence on $d$ in the convergence rate.

\vspace*{-3mm}

\section{Conclusion}
\label{sec:conc}
In this paper, we investigated sketched adaptive methods for FL. While the motivation behind combining sketching and adaptive methods for FL is clear, there is limited understanding on its empirical success due to the inherent technical challenges. We consider both mild-noise and heavy-tailed noise settings, propose corresponding adaptive algorithms for each, and show highly promising theoretical and empirical results. Inspired by the recently observations on heterogeneity in weights across neural network layers~\citep{zhang2024transformers}, an important future direction is to independently sketch layer-wise gradients, rather than sketching the concatenated gradient vectors. We believe our novel work can form the basis for future advances on the theme. 

\bibliographystyle{apalike}
\bibliography{arxiv/ref.bib}

\begin{thebibliography}{}

\bibitem[Alistarh et~al., 2017]{alistarh2017qsgd}
Alistarh, D., Grubic, D., Li, J., Tomioka, R., and Vojnovic, M. (2017).
\newblock Qsgd: Communication-efficient sgd via gradient quantization and encoding.
\newblock {\em Advances in neural information processing systems}, 30.

\bibitem[Alistarh et~al., 2018]{alistarh2018convergence}
Alistarh, D., Hoefler, T., Johansson, M., Konstantinov, N., Khirirat, S., and Renggli, C. (2018).
\newblock The convergence of sparsified gradient methods.
\newblock {\em Advances in Neural Information Processing Systems}, 31.

\bibitem[Basu et~al., 2019]{basu2019qsparse}
Basu, D., Data, D., Karakus, C., and Diggavi, S. (2019).
\newblock Qsparse-local-sgd: Distributed sgd with quantization, sparsification and local computations.
\newblock {\em Advances in Neural Information Processing Systems}, 32.

\bibitem[Bernstein et~al., 2018]{bernstein2018signsgd}
Bernstein, J., Wang, Y.-X., Azizzadenesheli, K., and Anandkumar, A. (2018).
\newblock signsgd: Compressed optimisation for non-convex problems.
\newblock In {\em International Conference on Machine Learning}, pages 560--569. PMLR.

\bibitem[Beznosikov et~al., 2023]{beznosikov2023biased}
Beznosikov, A., Horvath, S., Richtarik, P., and Safaryan, M. (2023).
\newblock On biased compression for distributed learning.
\newblock {\em Journal of Machine Learning Research}, 24(276):1--50.

\bibitem[Charikar et~al., 2002]{charikar2002finding}
Charikar, M., Chen, K., and Farach-Colton, M. (2002).
\newblock Finding frequent items in data streams.
\newblock In {\em International Colloquium on Automata, Languages, and Programming}, pages 693--703. Springer.

\bibitem[Chen et~al., 2021]{chen2021quantized}
Chen, C., Shen, L., Huang, H., and Liu, W. (2021).
\newblock Quantized adam with error feedback.
\newblock {\em ACM Transactions on Intelligent Systems and Technology (TIST)}, 12(5):1--26.

\bibitem[Chen et~al., 2022]{chen2022efficient}
Chen, C., Shen, L., Liu, W., and Luo, Z.-Q. (2022).
\newblock Efficient-adam: Communication-efficient distributed adam with complexity analysis.
\newblock {\em arXiv preprint arXiv:2205.14473}.

\bibitem[Chen et~al., 2023]{chen2023nqfl}
Chen, G., Xie, K., Tu, Y., Song, T., Xu, Y., Hu, J., and Xin, L. (2023).
\newblock Nqfl: Nonuniform quantization for communication efficient federated learning.
\newblock {\em IEEE Communications Letters}.

\bibitem[Chezhegov et~al., 2024]{chezhegov2024gradient}
Chezhegov, S., Klyukin, Y., Semenov, A., Beznosikov, A., Gasnikov, A., Horvath, S., Takac, M., and Gorbunov, E. (2024).
\newblock Gradient clipping improves adagrad when the noise is heavy-tailed.
\newblock {\em arXiv preprint arXiv:2406.04443}.

\bibitem[Cutkosky and Mehta, 2021]{cutkosky2021high}
Cutkosky, A. and Mehta, H. (2021).
\newblock High-probability bounds for non-convex stochastic optimization with heavy tails.
\newblock {\em Advances in Neural Information Processing Systems}, 34:4883--4895.

\bibitem[Devlin, 2018]{devlin2018bert}
Devlin, J. (2018).
\newblock Bert: Pre-training of deep bidirectional transformers for language understanding.
\newblock {\em arXiv preprint arXiv:1810.04805}.

\bibitem[Dosovitskiy et~al., 2020]{dosovitskiy2020image}
Dosovitskiy, A., Beyer, L., Kolesnikov, A., Weissenborn, D., Zhai, X., Unterthiner, T., Dehghani, M., Minderer, M., Heigold, G., Gelly, S., et~al. (2020).
\newblock An image is worth 16x16 words: Transformers for image recognition at scale.
\newblock {\em arXiv preprint arXiv:2010.11929}.

\bibitem[Duchi et~al., 2011]{duchi2011adaptive}
Duchi, J., Hazan, E., and Singer, Y. (2011).
\newblock Adaptive subgradient methods for online learning and stochastic optimization.
\newblock {\em Journal of machine learning research}, 12(7).

\bibitem[Fatkhullin et~al., 2024]{fatkhullin2024momentum}
Fatkhullin, I., Tyurin, A., and Richt{\'a}rik, P. (2024).
\newblock Momentum provably improves error feedback!
\newblock {\em Advances in Neural Information Processing Systems}, 36.

\bibitem[Ghorbani et~al., 2019]{ghorbani2019investigation}
Ghorbani, B., Krishnan, S., and Xiao, Y. (2019).
\newblock An investigation into neural net optimization via hessian eigenvalue density.
\newblock In {\em International Conference on Machine Learning}, pages 2232--2241. PMLR.

\bibitem[Gorbunov et~al., 2021a]{gorbunov2021marina}
Gorbunov, E., Burlachenko, K.~P., Li, Z., and Richt{\'a}rik, P. (2021a).
\newblock Marina: Faster non-convex distributed learning with compression.
\newblock In {\em International Conference on Machine Learning}, pages 3788--3798. PMLR.

\bibitem[Gorbunov et~al., 2020]{gorbunov2020stochastic}
Gorbunov, E., Danilova, M., and Gasnikov, A. (2020).
\newblock Stochastic optimization with heavy-tailed noise via accelerated gradient clipping.
\newblock {\em Advances in Neural Information Processing Systems}, 33:15042--15053.

\bibitem[Gorbunov et~al., 2021b]{gorbunov2021near}
Gorbunov, E., Danilova, M., Shibaev, I., Dvurechensky, P., and Gasnikov, A. (2021b).
\newblock Near-optimal high probability complexity bounds for non-smooth stochastic optimization with heavy-tailed noise.
\newblock {\em arXiv preprint arXiv:2106.05958}.

\bibitem[Gressmann et~al., 2020]{gressmann2020improving}
Gressmann, F., Eaton-Rosen, Z., and Luschi, C. (2020).
\newblock Improving neural network training in low dimensional random bases.
\newblock {\em Advances in Neural Information Processing Systems}, 33:12140--12150.

\bibitem[Haddadpour et~al., 2021]{haddadpour2021federated}
Haddadpour, F., Kamani, M.~M., Mokhtari, A., and Mahdavi, M. (2021).
\newblock Federated learning with compression: Unified analysis and sharp guarantees.
\newblock In {\em International Conference on Artificial Intelligence and Statistics}, pages 2350--2358. PMLR.

\bibitem[Harvey et~al., 2019]{harvey2019tight}
Harvey, N.~J., Liaw, C., Plan, Y., and Randhawa, S. (2019).
\newblock Tight analyses for non-smooth stochastic gradient descent.
\newblock In {\em Conference on Learning Theory}, pages 1579--1613. PMLR.

\bibitem[Ipsen and Saibaba, 2024]{ipsen2024stable}
Ipsen, I.~C. and Saibaba, A.~K. (2024).
\newblock Stable rank and intrinsic dimension of real and complex matrices.
\newblock {\em arXiv preprint arXiv:2407.21594}.

\bibitem[Ivkin et~al., 2019]{ivkin2019communication}
Ivkin, N., Rothchild, D., Ullah, E., Stoica, I., Arora, R., et~al. (2019).
\newblock Communication-efficient distributed sgd with sketching.
\newblock {\em Advances in Neural Information Processing Systems}, 32.

\bibitem[Jiang et~al., 2024]{jiang2024correlation}
Jiang, S., Sharma, P., and Joshi, G. (2024).
\newblock Correlation aware sparsified mean estimation using random projection.
\newblock {\em Advances in Neural Information Processing Systems}, 36.

\bibitem[Kingma and Ba, 2014]{kingma2014adam}
Kingma, D.~P. and Ba, J. (2014).
\newblock Adam: A method for stochastic optimization.
\newblock {\em arXiv preprint arXiv:1412.6980}.

\bibitem[Krizhevsky et~al., 2009]{krizhevsky2009learning}
Krizhevsky, A., Hinton, G., et~al. (2009).
\newblock Learning multiple layers of features from tiny images.

\bibitem[Li et~al., 2022a]{li20221}
Li, C., Awan, A.~A., Tang, H., Rajbhandari, S., and He, Y. (2022a).
\newblock 1-bit lamb: communication efficient large-scale large-batch training with lamb’s convergence speed.
\newblock In {\em 2022 IEEE 29th International Conference on High Performance Computing, Data, and Analytics (HiPC)}, pages 272--281. IEEE.

\bibitem[Li et~al., 2020]{li2020hessian}
Li, X., Gu, Q., Zhou, Y., Chen, T., and Banerjee, A. (2020).
\newblock Hessian based analysis of sgd for deep nets: Dynamics and generalization.
\newblock In {\em Proceedings of the 2020 SIAM International Conference on Data Mining}, pages 190--198. SIAM.

\bibitem[Li et~al., 2022b]{li2022distributed}
Li, X., Karimi, B., and Li, P. (2022b).
\newblock On distributed adaptive optimization with gradient compression.
\newblock {\em arXiv preprint arXiv:2205.05632}.

\bibitem[Li and Orabona, 2020]{li2020high}
Li, X. and Orabona, F. (2020).
\newblock A high probability analysis of adaptive sgd with momentum.
\newblock {\em arXiv preprint arXiv:2007.14294}.

\bibitem[Liao and Mahoney, 2021]{liao2021hessian}
Liao, Z. and Mahoney, M.~W. (2021).
\newblock Hessian eigenspectra of more realistic nonlinear models.
\newblock {\em Advances in Neural Information Processing Systems}, 34:20104--20117.

\bibitem[Lin et~al., 2017]{lin2017deep}
Lin, Y., Han, S., Mao, H., Wang, Y., and Dally, W.~J. (2017).
\newblock Deep gradient compression: Reducing the communication bandwidth for distributed training.
\newblock {\em arXiv preprint arXiv:1712.01887}.

\bibitem[Liu et~al., 2023a]{liu2023communication}
Liu, H., He, F., and Cao, G. (2023a).
\newblock Communication-efficient federated learning for heterogeneous edge devices based on adaptive gradient quantization.
\newblock In {\em IEEE INFOCOM 2023-IEEE Conference on Computer Communications}, pages 1--10. IEEE.

\bibitem[Liu et~al., 2023b]{liu2023sophia}
Liu, H., Li, Z., Hall, D., Liang, P., and Ma, T. (2023b).
\newblock Sophia: A scalable stochastic second-order optimizer for language model pre-training.
\newblock {\em arXiv preprint arXiv:2305.14342}.

\bibitem[Lu et~al., 2013]{lu2013faster}
Lu, Y., Dhillon, P., Foster, D.~P., and Ungar, L. (2013).
\newblock Faster ridge regression via the subsampled randomized hadamard transform.
\newblock {\em Advances in neural information processing systems}, 26.

\bibitem[Madden et~al., 2024]{madden2024high}
Madden, L., Dall'Anese, E., and Becker, S. (2024).
\newblock High probability convergence bounds for non-convex stochastic gradient descent with sub-weibull noise.
\newblock {\em Journal of Machine Learning Research}, 25(241):1--36.

\bibitem[Mishchenko et~al., 2022]{mishchenko2022proxskip}
Mishchenko, K., Malinovsky, G., Stich, S., and Richt{\'a}rik, P. (2022).
\newblock Proxskip: Yes! local gradient steps provably lead to communication acceleration! finally!
\newblock In {\em International Conference on Machine Learning}, pages 15750--15769. PMLR.

\bibitem[Mou et~al., 2020]{mou2020linear}
Mou, W., Li, C.~J., Wainwright, M.~J., Bartlett, P.~L., and Jordan, M.~I. (2020).
\newblock On linear stochastic approximation: Fine-grained polyak-ruppert and non-asymptotic concentration.
\newblock In {\em Conference on learning theory}, pages 2947--2997. PMLR.

\bibitem[Rabbani et~al., 2021]{rabbani2021comfetch}
Rabbani, T., Feng, B., Yang, Y., Rajkumar, A., Varshney, A., and Huang, F. (2021).
\newblock Comfetch: Federated learning of large networks on memory-constrained clients via sketching.
\newblock {\em arXiv e-prints}, pages arXiv--2109.

\bibitem[Rakhlin et~al., 2011]{rakhlin2011making}
Rakhlin, A., Shamir, O., and Sridharan, K. (2011).
\newblock Making gradient descent optimal for strongly convex stochastic optimization.
\newblock {\em arXiv preprint arXiv:1109.5647}.

\bibitem[Reddi et~al., 2020]{reddi2020adaptive}
Reddi, S., Charles, Z., Zaheer, M., Garrett, Z., Rush, K., Konecny, J., Kumar, S., and McMahan, H.~B. (2020).
\newblock Adaptive federated optimization.
\newblock {\em arXiv preprint arXiv:2003.00295}.

\bibitem[Reddi et~al., 2019]{reddi2019convergence}
Reddi, S.~J., Kale, S., and Kumar, S. (2019).
\newblock On the convergence of adam and beyond.
\newblock {\em arXiv preprint arXiv:1904.09237}.

\bibitem[Reisizadeh et~al., 2020]{reisizadeh2020fedpaq}
Reisizadeh, A., Mokhtari, A., Hassani, H., Jadbabaie, A., and Pedarsani, R. (2020).
\newblock Fedpaq: A communication-efficient federated learning method with periodic averaging and quantization.
\newblock In {\em International conference on artificial intelligence and statistics}, pages 2021--2031. PMLR.

\bibitem[Richt{\'a}rik et~al., 2024]{richtarik2024error}
Richt{\'a}rik, P., Gasanov, E., and Burlachenko, K. (2024).
\newblock Error feedback reloaded: From quadratic to arithmetic mean of smoothness constants.
\newblock {\em arXiv preprint arXiv:2402.10774}.

\bibitem[Richt{\'a}rik et~al., 2021]{richtarik2021ef21}
Richt{\'a}rik, P., Sokolov, I., and Fatkhullin, I. (2021).
\newblock Ef21: A new, simpler, theoretically better, and practically faster error feedback.
\newblock {\em Advances in Neural Information Processing Systems}, 34:4384--4396.

\bibitem[Richt{\'a}rik et~al., 2022]{richtarik20223pc}
Richt{\'a}rik, P., Sokolov, I., Gasanov, E., Fatkhullin, I., Li, Z., and Gorbunov, E. (2022).
\newblock 3pc: Three point compressors for communication-efficient distributed training and a better theory for lazy aggregation.
\newblock In {\em International Conference on Machine Learning}, pages 18596--18648. PMLR.

\bibitem[Rothchild et~al., 2020]{rothchild2020fetchsgd}
Rothchild, D., Panda, A., Ullah, E., Ivkin, N., Stoica, I., Braverman, V., Gonzalez, J., and Arora, R. (2020).
\newblock Fetchsgd: Communication-efficient federated learning with sketching.
\newblock In {\em International Conference on Machine Learning}, pages 8253--8265. PMLR.

\bibitem[Sadiev et~al., 2023]{sadiev2023high}
Sadiev, A., Danilova, M., Gorbunov, E., Horvath, S., Gidel, G., Dvurechensky, P., Gasnikov, A., and Richtarik, P. (2023).
\newblock High-probability bounds for stochastic optimization and variational inequalities: the case of unbounded variance.
\newblock In {\em International Conference on Machine Learning}, pages 29563--29648. PMLR.

\bibitem[Safaryan et~al., 2021]{safaryan2021fednl}
Safaryan, M., Islamov, R., Qian, X., and Richt{\'a}rik, P. (2021).
\newblock Fednl: Making newton-type methods applicable to federated learning.
\newblock {\em arXiv preprint arXiv:2106.02969}.

\bibitem[Sagun et~al., 2016]{sagun2016eigenvalues}
Sagun, L., Bottou, L., and LeCun, Y. (2016).
\newblock Eigenvalues of the hessian in deep learning: Singularity and beyond.
\newblock {\em arXiv preprint arXiv:1611.07476}.

\bibitem[Seide et~al., 2014]{seide20141}
Seide, F., Fu, H., Droppo, J., Li, G., and Yu, D. (2014).
\newblock 1-bit stochastic gradient descent and its application to data-parallel distributed training of speech dnns.
\newblock In {\em Interspeech}, volume 2014, pages 1058--1062. Singapore.

\bibitem[Shi et~al., 2019]{shi2019convergence}
Shi, S., Zhao, K., Wang, Q., Tang, Z., and Chu, X. (2019).
\newblock A convergence analysis of distributed sgd with communication-efficient gradient sparsification.
\newblock In {\em IJCAI}, pages 3411--3417.

\bibitem[Simsekli et~al., 2019]{simsekli2019tail}
Simsekli, U., Sagun, L., and Gurbuzbalaban, M. (2019).
\newblock A tail-index analysis of stochastic gradient noise in deep neural networks.
\newblock In {\em International Conference on Machine Learning}, pages 5827--5837. PMLR.

\bibitem[Sokolov and Richt{\'a}rik, 2024]{sokolov2024marina}
Sokolov, I. and Richt{\'a}rik, P. (2024).
\newblock Marina-p: Superior performance in non-smooth federated optimization with adaptive stepsizes.
\newblock {\em arXiv preprint arXiv:2412.17082}.

\bibitem[Song et~al., 2023]{song2023sketching}
Song, Z., Wang, Y., Yu, Z., and Zhang, L. (2023).
\newblock Sketching for first order method: efficient algorithm for low-bandwidth channel and vulnerability.
\newblock In {\em International Conference on Machine Learning}, pages 32365--32417. PMLR.

\bibitem[Spring et~al., 2019]{spring2019compressing}
Spring, R., Kyrillidis, A., Mohan, V., and Shrivastava, A. (2019).
\newblock Compressing gradient optimizers via count-sketches.
\newblock In {\em International Conference on Machine Learning}, pages 5946--5955. PMLR.

\bibitem[Stich, 2018]{stich2018local}
Stich, S.~U. (2018).
\newblock Local sgd converges fast and communicates little.
\newblock {\em arXiv preprint arXiv:1805.09767}.

\bibitem[Stich et~al., 2018]{stich2018sparsified}
Stich, S.~U., Cordonnier, J.-B., and Jaggi, M. (2018).
\newblock Sparsified sgd with memory.
\newblock {\em Advances in neural information processing systems}, 31.

\bibitem[Szlendak et~al., 2021]{szlendak2021permutation}
Szlendak, R., Tyurin, A., and Richt{\'a}rik, P. (2021).
\newblock Permutation compressors for provably faster distributed nonconvex optimization.
\newblock {\em arXiv preprint arXiv:2110.03300}.

\bibitem[Tang et~al., 2021]{tang20211}
Tang, H., Gan, S., Awan, A.~A., Rajbhandari, S., Li, C., Lian, X., Liu, J., Zhang, C., and He, Y. (2021).
\newblock 1-bit adam: Communication efficient large-scale training with adam’s convergence speed.
\newblock In {\em International Conference on Machine Learning}, pages 10118--10129. PMLR.

\bibitem[Tang et~al., 2024]{tang2024z}
Tang, Z., Wang, Y., and Chang, T.-H. (2024).
\newblock z-signfedavg: A unified stochastic sign-based compression for federated learning.
\newblock In {\em Proceedings of the AAAI Conference on Artificial Intelligence}, volume~38, pages 15301--15309.

\bibitem[Vargaftik et~al., 2021]{vargaftik2021drive}
Vargaftik, S., Ben-Basat, R., Portnoy, A., Mendelson, G., Ben-Itzhak, Y., and Mitzenmacher, M. (2021).
\newblock Drive: One-bit distributed mean estimation.
\newblock {\em Advances in Neural Information Processing Systems}, 34:362--377.

\bibitem[Vershynin, 2018]{vershynin2018high}
Vershynin, R. (2018).
\newblock {\em High-dimensional probability: An introduction with applications in data science}, volume~47.
\newblock Cambridge university press.

\bibitem[Vladimirova et~al., 2020]{vladimirova2020sub}
Vladimirova, M., Girard, S., Nguyen, H., and Arbel, J. (2020).
\newblock Sub-weibull distributions: Generalizing sub-gaussian and sub-exponential properties to heavier tailed distributions.
\newblock {\em Stat}, 9(1):e318.

\bibitem[Wang et~al., 2018]{wang2018glue}
Wang, A., Singh, A., Michael, J., Hill, F., Levy, O., and Bowman, S.~R. (2018).
\newblock Glue: A multi-task benchmark and analysis platform for natural language understanding.
\newblock {\em arXiv preprint arXiv:1804.07461}.

\bibitem[Wang et~al., 2021]{wang2021error}
Wang, H., Guo, S., Qu, Z., Li, R., and Liu, Z. (2021).
\newblock Error-compensated sparsification for communication-efficient decentralized training in edge environment.
\newblock {\em IEEE Transactions on Parallel and Distributed Systems}, 33(1):14--25.

\bibitem[Wang and Joshi, 2019]{wang2019adaptive}
Wang, J. and Joshi, G. (2019).
\newblock Adaptive communication strategies to achieve the best error-runtime trade-off in local-update sgd.
\newblock {\em Proceedings of Machine Learning and Systems}, 1:212--229.

\bibitem[Wang et~al., 2023]{wang2023cocktailsgd}
Wang, J., Lu, Y., Yuan, B., Chen, B., Liang, P., De~Sa, C., Re, C., and Zhang, C. (2023).
\newblock Cocktailsgd: Fine-tuning foundation models over 500mbps networks.
\newblock In {\em International Conference on Machine Learning}, pages 36058--36076. PMLR.

\bibitem[Wang et~al., 2022]{wang2022communication}
Wang, Y., Lin, L., and Chen, J. (2022).
\newblock Communication-compressed adaptive gradient method for distributed nonconvex optimization.
\newblock In {\em International Conference on Artificial Intelligence and Statistics}, pages 6292--6320. PMLR.

\bibitem[Wangni et~al., 2018]{wangni2018gradient}
Wangni, J., Wang, J., Liu, J., and Zhang, T. (2018).
\newblock Gradient sparsification for communication-efficient distributed optimization.
\newblock {\em Advances in Neural Information Processing Systems}, 31.

\bibitem[Wortsman et~al., 2021]{wortsman2021learning}
Wortsman, M., Horton, M.~C., Guestrin, C., Farhadi, A., and Rastegari, M. (2021).
\newblock Learning neural network subspaces.
\newblock In {\em International Conference on Machine Learning}, pages 11217--11227. PMLR.

\bibitem[Wu et~al., 2018]{wu2018error}
Wu, J., Huang, W., Huang, J., and Zhang, T. (2018).
\newblock Error compensated quantized sgd and its applications to large-scale distributed optimization.
\newblock In {\em International Conference on Machine Learning}, pages 5325--5333. PMLR.

\bibitem[Wu and He, 2018]{wu2018group}
Wu, Y. and He, K. (2018).
\newblock Group normalization.
\newblock In {\em Proceedings of the European conference on computer vision (ECCV)}, pages 3--19.

\bibitem[Xie et~al., 2020]{xie2020cser}
Xie, C., Zheng, S., Koyejo, S., Gupta, I., Li, M., and Lin, H. (2020).
\newblock Cser: Communication-efficient sgd with error reset.
\newblock {\em Advances in Neural Information Processing Systems}, 33:12593--12603.

\bibitem[Xie et~al., 2022]{xie2022power}
Xie, Z., Tang, Q.-Y., Cai, Y., Sun, M., and Li, P. (2022).
\newblock On the power-law hessian spectrums in deep learning.
\newblock {\em arXiv preprint arXiv:2201.13011}.

\bibitem[Xu et~al., 2023]{xu2023slamb}
Xu, H., Zhang, W., Fei, J., Wu, Y., Xie, T., Huang, J., Xie, Y., Elhoseiny, M., and Kalnis, P. (2023).
\newblock Slamb: accelerated large batch training with sparse communication.
\newblock In {\em International Conference on Machine Learning}, pages 38801--38825. PMLR.

\bibitem[Yang et~al., 2022]{yang2022taming}
Yang, H., Qiu, P., and Liu, J. (2022).
\newblock Taming fat-tailed (“heavier-tailed” with potentially infinite variance) noise in federated learning.
\newblock {\em Advances in Neural Information Processing Systems}, 35:17017--17029.

\bibitem[Yao et~al., 2020]{yao2020pyhessian}
Yao, Z., Gholami, A., Keutzer, K., and Mahoney, M.~W. (2020).
\newblock Pyhessian: Neural networks through the lens of the hessian.
\newblock In {\em 2020 IEEE international conference on big data (Big data)}, pages 581--590. IEEE.

\bibitem[Zhang et~al., 2020]{zhang2020adaptive}
Zhang, J., Karimireddy, S.~P., Veit, A., Kim, S., Reddi, S., Kumar, S., and Sra, S. (2020).
\newblock Why are adaptive methods good for attention models?
\newblock {\em Advances in Neural Information Processing Systems}, 33:15383--15393.

\bibitem[Zhang et~al., 2024]{zhang2024transformers}
Zhang, Y., Chen, C., Ding, T., Li, Z., Sun, R., and Luo, Z.-Q. (2024).
\newblock Why transformers need adam: A hessian perspective.
\newblock {\em arXiv preprint arXiv:2402.16788}.

\bibitem[Zheng et~al., 2019]{zheng2019communication}
Zheng, S., Huang, Z., and Kwok, J. (2019).
\newblock Communication-efficient distributed blockwise momentum sgd with error-feedback.
\newblock {\em Advances in Neural Information Processing Systems}, 32.

\end{thebibliography}


\appendix
\newpage

\section{Lemma for Random Sketching}
\label{app:sketch}
For completeness, we provide the following lemmas that give high probability bounds on the inner products.
\begin{lemma} (SRHT)[Same as Lemma D.23~\cite{song2023sketching}]
    Let $R \in \R^{b\times d}$ denote a subsample randomized Hadamard transform or AMS sketching matrix. Then for any fixed vector $h \in \R$ and any fixed vector $g \in \R$ the following properties hold:
    \begin{align*}
        \P\left[\vert \langle g^\top R^\top R h - g^\top h\vert \ge \logd \norm{g}_2\norm{h}_2 \right] \le \Theta(\delta).
    \end{align*}
    \label{lemma_srht}
\end{lemma}

\begin{lemma} (Gaussian)[Same as Lemma D.24~\cite{song2023sketching}]
    Let $R \in \R^{b\times d}$ denote a random Gaussian matrix. Then for any fixed vector $h \in \R$ and any fixed vector $g \in \R$ the following properties hold:
    \begin{align*}
        \P\left[\vert \langle g^\top R^\top R h - g^\top h\vert \ge \logd \norm{g}_2\norm{h}_2 \right] \le \Theta(\delta).
    \end{align*}
    \label{lemma_gaussian}
\end{lemma}

\begin{lemma} (Count-Sketch)[Same as Lemma D.25~\cite{song2023sketching}]
    Let $R \in \R^{b\times d}$ denote a count-sketch matrix. Then for any fixed vector $h \in \R$ and any fixed vector $g \in \R$ the following properties hold:
    \begin{align*}
        \P\left[\vert \langle g^\top R^\top R h - g^\top h\vert \ge \log(1/\delta) \norm{g}_2\norm{h}_2 \right] \le \Theta(\delta).
    \end{align*}
    \label{lemma_count_sketch}
\end{lemma}

\section{Proof of Theorem~\ref{theorem_mild}}
\label{app:mild}
\subsection{Proof of Lemma~\ref{lemma_descent_amsgrad}}
Let
\begin{align*}
z_t = x_t + \frac{\beta_1}{1-\beta_1}(x_t - x_{t-1}) = \frac{1}{1-\beta_1}x_t - \frac{\beta_1}{1-\beta_1}x_{t-1}.
\end{align*}
Then, the update on $z_t$ can be expressed as
\begin{align*}
    z_{t+1}-z_t &= \frac{1}{1-\beta_1}(x_{t+1}-x_t) - \frac{\beta_1}{1-\beta_1}(x_t-x_{t-1}) \\
    &= -\frac{1}{1-\beta_1}\lr \hat{V_t}^{-1/2} \cdot m_t + \frac{\beta_1}{1-\beta_1}\lr \hat{V_{t-1}}^{-1/2}\cdot m_{t-1} \\
    &= -\frac{1}{1-\beta_1}\lr \hat{V_t}^{-1/2} \cdot(\beta_1 m_{t-1} + (1-\beta_1)\cdot R_t^\top \barm_t) + \frac{\beta_1}{1-\beta_1}\lr {\hat{V}_{t-1}}^{-1/2}\cdot m_{t-1} \\
    &= \frac{\beta_1}{1-\beta_1}\left(\lr {\hat{V}_{t-1}}^{-1/2}- \lr \hat{V_t}^{-1/2}\right)m_{t-1} - \frac{\lr}{C} \hat{V_t}^{-1/2} R_t^\top \sum_{c=1}^C \barm_{t}^c \\
    &= \frac{\beta_1}{1-\beta_1}\left(\lr {\hat{V}_{t-1}}^{-1/2}- \lr \hat{V_t}^{-1/2}\right)m_{t-1} - \frac{\lr}{C} \hat{V_t}^{-1/2} R_t^\top \sum_{c=1}^C R_t (x_{t,0}^c - x_{t,K}^c) \\
    &= \frac{\beta_1}{1-\beta_1}\left(\lr {\hat{V}_{t-1}}^{-1/2}- \lr \hat{V_t}^{-1/2}\right)m_{t-1} - \frac{\lr \etat}{C} \hat{V_t}^{-1/2} \sum_{c=1}^C  \sum_{k=1}^K R_t^\top R_t g_{t,k}^c 
\end{align*}

By Taylor expansion, we have
\begin{align}
    \cL(z_{t+1}) &= \cL(z_t) + \nabla \cL(z_t)^\top (z_{t+1}-z_t) + \frac{1}{2}(z_{t+1}-z_t)^\top \hat H_{\cL} (z_{t+1}-z_t) \notag \\
    &= \cL(z_t) + \nabla \cL(x_t)^\top (z_{t+1}-z_t) +  (\nabla \cL(z_t) - \nabla \cL(x_t))^\top (z_{t+1}-z_t) +\frac{1}{2} (z_{t+1}-z_t)^\top \hat H_{\cL} (z_{t+1}-z_t). 
\end{align}

Bounding the first-order term
\begin{align*}
&\nabla \cL(x_t)^\top (z_{t+1}-z_t) \\
=& \nabla \cL(x_t)^\top \left(  \frac{\beta_1}{1-\beta_1}\left(\lr {\hat{V}_{t-1}}^{-1/2}- \lr \hat{V_t}^{-1/2}\right)m_{t-1} - \frac{\lr\etat}{C} \hat{V_t}^{-1/2} \sum_{c=1}^C  \sum_{k=1}^K R_t^\top R_t g_{t,k}^c \right)\\
\le &  \frac{\beta_1}{1-\beta_1}  \cL(x_t)^\top \left(\lr {\hat{V}_{t-1}}^{-1/2}- \lr \hat{V_t}^{-1/2}\right) m_{t-1} -  \frac{\etat}{C}\nabla \cL(x_t)^\top (\lr \hat{V_t}^{-1/2}-\lr {\hat{V}_{t-1}}^{-1/2})  \sum_{c=1}^C  \sum_{k=1}^K R_t^\top R_t g_{t,k}^c \\
&  - \frac{\lr\etat}{C}\nabla \cL(x_t)^\top {\hat{V}_{t-1}}^{-1/2}  \sum_{c=1}^C  \sum_{k=1}^K R_t^\top R_t g_{t,k}^c 
\end{align*}

For the difference term, applying Lemma~\ref{lemma_gaussian} yields
\begin{align*}
    &\frac{\etat}{C}\nabla \cL(x_t)^\top (\lr \hat{V_t}^{-1/2}-\lr {\hat{V}_{t-1}}^{-1/2})  \sum_{c=1}^C  \sum_{k=1}^K R_t^\top R_t g_{t,k}^c \\
    \le& \frac{\etat \lr}{C}(1+\logtd) \Vert \nabla \cL(x_t) \Vert \Vert \hat{V_t}^{-1/2}-{\hat{V}_{t-1}}^{-1/2} \Vert_2 \sum_{c=1}^C  \sum_{k=1}^K\Vert g_{t,k}^c \Vert
\end{align*}
Denote $[\cdot]_i$ as the $i$-th element of a vector. The $l2$-norm
\begin{align*}
    \Vert \hat{V_t}^{-1/2}-{\hat{V}_{t-1}}^{-1/2} \Vert_2 & = \max_i \frac{1}{\sqrt{\hat{v}_{t-1, i}}+\epsilon} - \frac{1}{\sqrt{\hat{v}_{t, i}}+\epsilon} = \max_i \frac{\sqrt{\hat{v}_{t, i}} - \sqrt{\hat{v}_{t-1, i}}}{(\sqrt{\hat{v}_{t-1, i}}+\epsilon)(\sqrt{\hat{v}_{t, i}}+\epsilon)} \\
    & = \max_i \frac{\hat{v}_{t, i} - \hat{v}_{t-1, i}}{(\sqrt{\hat{v}_{t-1, i}}+\epsilon)(\sqrt{\hat{v}_{t, i}}+\epsilon)(\sqrt{\hat{v}_{t, i}} + \sqrt{\hat{v}_{t-1, i}})} 
\end{align*}
By definition, $\hat{v}_{t} = \max(\hat{v}_{t-1}, v_t)$. If $\hat{v}_{t, i} = \hat{v}_{t-1, i}$, the RHS is 0. Otherwise, $\hat{v}_{t,i} = v_{t,i}$.
\begin{align*}
     \Vert \hat{V_t}^{-1/2}-{\hat{V}_{t-1}}^{-1/2} \Vert_2 &\le \max_i \frac{v_{t,i} - v_{t-1,i}}{(\sqrt{\hat{v}_{t-1, i}}+\epsilon)(\sqrt{\hat{v}_{t, i}}+\epsilon)(\sqrt{\hat{v}_{t, i}} + \sqrt{\hat{v}_{t-1, i}})} \\
     &\le \max_i \frac{(1-\beta_2)(\barv_{t,i} - v_{t-1,i})}{\epsilon^2 \sqrt{(1-\beta_2)\barv_{t,i}}} \\
     &\le \max_i \frac{\sqrt{1-\beta_2}}{\epsilon^2} \sqrt{\barv_{t,i}} \\
     &= \frac{\sqrt{1-\beta_2}}{\epsilon^2} \max_i \sqrt{\frac{\etat^2}{C}\sum_{c=1}^C [(\sum_{k=1}^K R_t^\top R_t g_{t,k}^c)^2]_i} \\
     &\le  \frac{2\etat \sqrt{(1-\beta_2)}}{\epsilon^2} \sqrt{1+\logdsqt}G.
\end{align*}
The first inequality is from $\hat{v}_{t-1,i} \ge v_{t-1,i}$. The second inequality comes from $\hat{v}_{t,i} \ge v_{t,i} \ge (1-\beta_2)\barv_{t,i}$. The last inequality follows from applying Lemma~\ref{lemma_gaussian} to each dimension of $g_{t,k}^c$. Plugging into the bound for the difference term
\begin{align*}
&\frac{\etat}{C}\nabla \cL(x_t)^\top (\lr \hat{V_t}^{-1/2}-\lr {\hat{V}_{t-1}}^{-1/2})  \sum_{c=1}^C  \sum_{k=1}^K R_t^\top R_t g_{t,k}^c \\
\le & \frac{2\etat^2 \lr  \sqrt{(1-\beta_2)}}{\epsilon^2} (1+\logdsqt)^{3/2} G^3
\end{align*}

The quadratic terms can be written as 
\begin{align*}
    &(\nabla \cL(z_t) - \nabla \cL(x_t))^\top (z_{t+1}-z_t) = (z_t - x_t)^\top \hat{H}_{\cL}(\frac{1}{1-\beta_1}(x_{t+1}-x_t) - \frac{\beta_1}{1-\beta_1}(x_t-x_{t-1})),
\end{align*}
where $\hat{H}_\cL$ is a second-order Taylor remainder. So the quadratic term can be further seen as a quadratic form over $z_{t+1}-z_t$ and $z_t - x_t$, denote as $\calQ(z_{t+1}-z_t, z_t - x_t)$. For the same reason, the term $\frac{1}{2} (z_{t+1}-z_t)^\top \hat H_{\cL} (z_{t+1}-z_t)$ can also be written into a quadratic form $\calQ(z_{t+1}-z_t, z_{t+1}-z_t)$. Putting the two terms together yields a quadratic form of $\calQ(z_{t+1}-z_t, z_t - x_t)$.

\subsection{Proof of Lemma~\ref{lemma_inner_v} (Generalized version of Lemma~\ref{lemma_inner_product})}
\begin{proof}
We can prove by induction. For $t=0$, since $m_0=0$, the inequality holds. Suppose we have for $h \in \mathbb{R}^d$, s.t. $\Vert h\Vert \le H$, with probability $1-\Theta((t-1)\delta)$,
\begin{align*}
    \vert m_{t-1}^\top h \vert \le (1+\frac{\log^{1.5}(CKd/\delta)}{\sqrt{b}}) G
\end{align*}
Then by the update rule,
\begin{align*}
\vert m_t^\top h \vert &= \vert (\beta_1 \cdot m_{t-1} + (1 - \beta_1) \cdot \frac{\etat}{C} \sum_{c=1}^C  \sum_{k=1}^K R_t^\top R_t g_{t,k}^c )^\top h\vert \\
& \le \beta_1 \vert m_{t-1}^\top h \vert + \frac{(1-\beta_1)\eta}{C}  \sum_{c=1}^C  \sum_{k=1}^K \vert \langle R_t^\top R_t g_{t,k}^c,h \rangle \vert \\
& \le \beta_1 \vert m_{t-1}^\top h \vert + (1-\beta_1) (1+\frac{\log^{1.5}(CKd/\delta)}{\sqrt{b}})  \eta  \sum_{k=1}^K \Vert g_{t,k}^c \Vert_2 \Vert h \Vert_2 \\
& \le (1+\frac{\log^{1.5}(CKd/\delta)}{\sqrt{b}}) \eta KGH,~~w.p.~ 1-\Theta(t \delta).
\end{align*}
Let $h=\hV v_i$. Then $\Vert h\Vert_2 \le 1/\epsilon$. We have
\begin{align*}
\vert (\hV m_{t})^\top v_i \vert &\le (1+\frac{\log^{1.5}(CKd/\delta)}{\sqrt{b}}) \eta K G / \epsilon
\end{align*}
\end{proof}

\subsection{Proof of Lemma~\ref{lemma_descent_bound}}
\label{sec:proof_lemma_descent_bound}
We first prove the element-wise lower bound of the diagonal matrix $\kV$. Denote $(\kV)_i$ as the $i$-th element on the diagonal of  $\kV$. By the update rule,
\begin{align*}
    (\kV)_i \ge (\max_{t-1}(\sqrt{v_{t,i}})+ \epsilon)^{-1} \ge  (\sqrt{1+\logstd}\eta KG + \epsilon)^{-1}, ~~w.p.~ 1 -\Theta(\delta)
\end{align*}
where the last inequality follows by letting $h$ as a one-hot vector $h_i$ in Lemma~\ref{lemma_srht}, observing that the elements can be transformed to an inner product form $v_{t, i} = v_t^\top h_i$. Then the scaled gradient norm can be lower bounded as
\begin{align*}
\nabla \cL(x_t)^\top {\hat{V}_{t-1}}^{-1/2} \nabla \cL(x_t) 
& \ge  \min_i(\kV)_i \sum_{i=1}^d  [\nabla \cL(x_t)]_i^2 \\
&\ge (\sqrt{1+\logstd}\eta KG + \epsilon)^{-1} \Vert \nabla \cL (x_t) \Vert^2, ~~w.p.~ 1 -\Theta(d\delta)
\end{align*}
which completes the proof by applying union bounded on the dimension $d$.

\subsection{Proof of Lemma~\ref{lemma_martingale}}
Since the noise is zero-centered, we view the random process of 
\begin{align*}
    \{Y_t = \sum_{\tau=1}^t\sumv{C}\sum_{k=1}^K \nabla \cL(x_\tau)^\top {\hat{V}_{\tau-1}}^{-1/2} (R_\tau^\top R_\tau g_{\tau,k}^c - g_{\tau,k}^c)\}_{t=1}^T
\end{align*}
 as a martingale. The difference of $\vert Y_{t+1} - Y_t\vert$ is bounded with high probability
\begin{align*}
    \vert Y_{t+1} - Y_t\vert = \vert  \nabla \cL(x_t)^\top {\hat{V}_{t-1}}^{-1/2} (R_t^\top R_t g_{t,k}^c - g_{t,k}^c) \vert \le \frac{\log^{1.5}(d/\delta)}{\sqrt{b}} G \Vert \hV  \nabla \cL(x_t)\Vert_2, ~~w.p.~ 1-\Theta(\delta)
\end{align*}

Then by Azuma's inequality,
\begin{align*}
\P(\vert Y_T \vert \ge \martin \sqrt{\sum_{t=1}^T \left( \frac{\log^{1.5}(d/\delta)}{\sqrt{b}} G \Vert \hV  \nabla \cL(x_t)\Vert_2\right)^2} ) = O({\rm exp}(-\Omega(\martin^2))) + T \delta
\end{align*}
Note that the original Azuma's is conditioned on a uniform bound of the difference term, while our bound here is of high probability. Hence, we need another union bound. A similar bound can be achieved for the sub-Gaussian noise in stochastic gradient. Let 
\begin{align*}
    Z_t =  \sum_{\tau=1}^t \sumv{C} \sum_{k=1}^K  \nabla \cL(x_\tau)^\top {\hat{V}_{\tau-1}}^{-1/2} ( g_{\tau,k}^c - \nabla \cL^c(x_{t,k}^c)).
\end{align*}
Then
\begin{align*}
    \P(\vert Z_T \vert \ge \martin \sqrt{\sum_{t=1}^T \frac{\sigma^2}{\epsilon^2} \log(\frac{2T}{\delta_g}) } ) = O({\rm exp}(-\Omega(\martin^2))) + \delta_g
\end{align*}
Combining the two bounds by union bound completes the proof.

\subsection{Proof of Theorem~\ref{theorem_mild}}
We first introduce the lemma 
\begin{lemma}
\label{lemma_descent_amsgrad} (Informal Version of Lemma~\ref{lemma_descent_amsgrad})
    For any round $t \in [T]$, 
    \begin{small}
    \begin{align*}
    \cL(z_{t+1}) \lessapprox &\cL(z_t) - \frac{\kappa\etat}{C}\sum_{c=1}^C  \sum_{k=1}^K\nabla \cL(x_t)^\top {\hat{V}_{t-1}}^{-1/2}   R_t^\top R_t g_{t,k}^c +  (z_t - x_t)^\top H_{\cL}(\hat{z}_t)(z_{t+1} - z_t) + \frac{2\etat^2 \lr}{(1-\beta_1)\epsilon^2} (1+\logdsqt)^{3/2} G^3,
    \end{align*}
    \end{small}
    where $ H_{\cL}(\hat{z}_t)$ is the loss Hessian at some $\hat{z}_t$ within the element-wise interval of $[x_t, z_t]$, and $\lessapprox$ omits the less important terms.
\end{lemma}

After applying Lemma~\ref{lemma_descent_amsgrad}. The second order quadratic forms in the descent lemma can be written as
\begin{align*}
    &(\nabla \cL(z_t) - \nabla \cL(x_t))^\top (z_{t+1}-z_t) \\
    =& (z_t - x_t)^\top \hat{H}_{\cL}(\frac{1}{1-\beta_1}(x_{t+1}-x_t) - \frac{\beta_1}{1-\beta_1}(x_t-x_{t-1})) \\
    =& -\lr\frac{\beta_1}{1-\beta_1} (\kV m_{t-1})^\top \hat{H}_{\cL} (\frac{1}{1-\beta_1}(-\lr \hV m_t) - \frac{\beta_1}{1-\beta_1}(-\lr \kV m_{t-1})) \\
    =& \lr^2 \frac{\beta_1}{(1-\beta_1)^2}(\kV m_{t-1})^\top \hat{H}_{\cL} (\hV m_t) - \lr^2 \frac{\beta_1^2}{(1-\beta_1)^2} (\kV m_{t-1})^\top \hat{H}_{\cL} (\kV m_{t-1}),
\end{align*}
and 
\begin{align*}
    &(z_{t+1} - z_t)^\top \hat{H}_\cL (z_{t+1}-z_t) \\
    =& (\frac{1}{1-\beta_1}(x_{t+1}-x_t) - \frac{\beta_1}{1-\beta_1}(x_t-x_{t-1}))^\top \hat{H}_{\cL}(\frac{1}{1-\beta_1}(x_{t+1}-x_t) - \frac{\beta_1}{1-\beta_1}(x_t-x_{t-1})) \\
    =& \frac{1}{(1-\beta_1)^2}(x_{t+1}-x_t)^\top  \hat{H}_{\cL} (x_{t+1}-x_t) - \frac{2\beta_1}{(1-\beta_1)^2}(x_{t+1}-x_t)^\top  \hat{H}_{\cL} (x_{t}-x_{t-1})\\
    & + \frac{\beta_1^2}{(1-\beta_1)^2}(x_{t}-x_{t-1})^\top  \hat{H}_{\cL} (x_{t}-x_{t-1}),
\end{align*}
which is essentially a quadratic form defined on $\hV m_t$ and $\kV m_{t-1}$. Hence, we provide a generalized version of Lemma~\ref{lemma_inner_product}, as follows.
\begin{lemma}
\label{lemma_inner_v}
With probability $1-\Theta(t \delta)$, for eigenvector $v_i$ of the Hessian matrix, $\vert (\hV m_{t})^\top v_i \vert \le (1+\frac{\log^{1.5}(CKd/\delta)}{\sqrt{b}}) \eta K G / \epsilon$.
\end{lemma}
Note that  $v_i$ can be any basis and is constant throughout the training process. Then the sum of quadratic forms is written as
\begin{align*}
    &(\nabla \cL(z_t) - \nabla \cL(x_t))^\top (z_{t+1}-z_t) \\
    \le & \lr^2 \frac{\beta_1}{(1-\beta_1)^2}(\kV m_{t-1})^\top \hat{H}_{\cL} (\hV m_t) - \lr^2 \frac{\beta_1^2}{(1-\beta_1)^2} (\kV m_{t-1})^\top \hat{H}_{\cL} (\kV m_{t-1}), \\
    =& \lr^2 \frac{\beta_1}{(1-\beta_1)^2}\sum_{i=1}^d \lambda_i  (\kV m_{t-1})^\top (v_i v_i^\top)\hV m_t - \lr^2 \frac{\beta_1^2}{(1-\beta_1)^2} \sum_{i=1}^d \lambda_i  (\kV m_{t-1})^\top (v_i v_i^\top) \kV m_{t-1} \\
    \le & \lr^2 \frac{\beta_1}{(1-\beta_1)^2}\sum_{i=1}^d \vert \lambda_i\vert  \vert(\kV m_{t-1})^\top v_i\vert \vert(\hV m_t)^\top v_i \vert + \lr^2 \frac{\beta_1^2}{(1-\beta_1)^2} \sum_{i=1}^d \vert\lambda_i\vert  \vert (\kV m_{t-1})^\top v_i \vert^2 \\
    \le &  \lr^2 \frac{2}{(1-\beta_1)^2} \intdim L (1+\frac{\log^{1.5}(CKd/\delta)}{\sqrt{b}})^2 \eta^2 K^2 G^2 / \epsilon^2,
\end{align*}
where the last inequality is by $\beta_1 \le 1$ and Lemma.~\ref{lemma_inner_v}.

\textbf{First-Order Descent Term}. 
The first-order term in the descent lemma can be decomposed into three components, which we will handle separately.
\begin{align*}
    \nabla \cL(x_t)^\top {\hat{V}_{t-1}}^{-1/2}   R_t^\top R_t g_{t,k}^c =& \underbrace{\nabla \cL(x_t)^\top {\hat{V}_{t-1}}^{-1/2} \nabla \cL^c(x_t)}_{\cD_1}  +   \underbrace{\nabla \cL(x_t)^\top {\hat{V}_{t-1}}^{-1/2} (R_t^\top R_t g_{t,k}^c - \nabla \cL^c(x_{t,k}^c))}_{\cD_2} \\
    &+  \underbrace{\nabla \cL(x_t)^\top {\hat{V}_{t-1}}^{-1/2}(\nabla \cL^c(x_{t,k}^c) - \nabla \cL^c(x_t))}_{\cD_3}.
\end{align*}
First, $\cD_3$ 
can be reduced to a second-order term by smoothness over $\cL$,
\begin{align*}
    &\nabla \cL(x_t)^\top {\hat{V}_{t-1}}^{-1/2}(\nabla \cL^c(x_{t,k}^c) - \nabla \cL^c(x_t))
    =  \nabla \cL(x_t)^\top {\hat{V}_{t-1}}^{-1/2}\hH_L^c(x_{t,k}^c - x_t) \\
    = & -\eta \sum_{\tau=1}^k \nabla \cL(x_t)^\top {\hat{V}_{t-1}}^{-1/2}\hH_\cL^c g_{t, \tau}^c \\
    \le & \frac{1}{\epsilon} L \Vert \nabla L \Vert \sum_{\tau=1}^k\norm{g_{t, \tau}^c} \le \frac{1}{\epsilon} \eta LKG^2.
\end{align*}

Note that this term does not involve any stochasticity with regard to random sketching, which means we can directly derive the upper bound by Cauchy-Schwartz in the last inequality.

Next observing that $\sumv{C} \nabla \cL^c(x_t) = \nabla \cL(x_t)$, $\cD_1$ composes a scaled squared gradient norm. Applying element-wise high probability bound on random sketching yields the lower bound for the scale. 
By Lemma~\ref{lemma_descent_bound}, we can derive the lower bound for $\cD_1$. Note that applying union bound to $\cD_1$ does not introduce another $T$ dependence, since $\hat v_{t,i}$ is monotonically non-decreasing.

\textbf{Martingale for zero-centered noise.} $\cD_2$ contains a zero-centered noise term $R_t^\top R_t g_{t,k}^c - \nabla \cL^c(x_{t,k}^c)$, where the randomness is over $R_t$ and the mini-batch noise at round $t$. Despite $x_{t,k}^c$ has temporal dependence, the fresh noise at round $t$ is independent of the randomness in the previous iterations. Hence, the random process defined by the aggregation of these norm terms over time forms a martingale. By Lemma~\ref{lemma_martingale}, we can bound this term $\cD_2$.

Finally, putting these parts together by union bound over $[T]$ and telescoping  the descent lemma leads to Theorem~\ref{theorem_mild}.

\subsection{Proof of Corollary~\ref{corollary_mild}}
In the aysmptotic regime, with sufficiently large $T$, the term $\sqrt{1+\logtdsq} \eta K G$ approaches $\epsilon$, so the denominator on the LHS can be replaced with $2\epsilon$. Then the derivation is straightforward by just substituting $\eta = \frac{1}{\sqrt{T}K}$ into Theorem~\ref{theorem_mild}.

\subsection{Proof of Corollary~\ref{corollary_mild_near_init}}
We first develop the convergence bound in Theorem~\ref{theorem_mild} under the condition $b\ge \log^3(CKd^2T^2/\delta)$,
\begin{align*}
    \left(\sqrt{2} \eta K G + \epsilon \right)^{-1} \lr \eta K &\sum_{t=1}^T \Vert \nabla \cL (x_t) \Vert^2  
    \leq \cL(z_1) + \frac{1}{\epsilon} \lr \eta^2 L K^2G^2 T \\ &+ \martin \lr \eta K\sqrt{T}(\frac{G^2}{\epsilon} + \frac{\sigma}{\epsilon}\log^{\frac{1}{2}}(\frac{2T}{\delta_g})) 
    + \eta^2 \kappa^2 T \frac{32}{(1-\beta_1)^2} \frac{\intdim L K^2 G^2}{\epsilon^2},
\end{align*}
The condition on $T \le \frac{J_1-\sqrt{2}G}{\epsilon^2}$ is equivalent to
\begin{align*}
    \frac{\sqrt{2} \eta K G + \epsilon}{\eta K} \le J_1,
\end{align*}
since $\eta = \frac{1}{\sqrt{T}K}$. Then scaling the coefficient on the left hand side and substituting $\frac{1}{\sqrt{T}K}$ for $\eta$, we derive
\begin{align*}
    \frac{1}{J_1T}\sum_{t=1}^T \Vert \nabla \cL (x_t) \Vert^2  
    &\leq \frac{\cL(z_1)\epsilon}{\kappa T} + \frac{1}{\epsilon}  \frac{L G^2}{T}  + \frac{\martin }{T}(G^2 + \sigma\log^{\frac{1}{2}}(\frac{2T}{\delta_g})) +  \frac{\kappa}{T}\frac{32}{(1-\beta_1)^2} \frac{\intdim L G^2}{\epsilon},
    \end{align*}   

\subsection{A non-asymptotic bound on practical learning rates}
We first state a convergence bound on using practical learning rates, which decays as the optimization procedure.
\begin{theorem}
    Suppose the sequence of iterates $\{x_t\}_{t=1}^T$ is generated by Algorithm~\ref{alg:sketch_federated} with a decaying learning rate $\eta_t = \frac{1}{\sqrt{t+T_0}K}$, where $T_0=\lceil \frac{1}{1-\beta_1^2}\rceil$. Under Assumptions 1-4, for any $T$ and $\epsilon > 0$, with probability $1 - \Theta(\delta) - O(\exp(-\Omega(\martin^2))) -\delta_g $, 
    \begin{small}
    \begin{align*}
    &\sum_{t=1}^T \left(\sqrt{1+\logtdsq} \eta_t J K G + \epsilon \right)^{-1} \lr \eta_t \Vert \nabla \cL (x_t) \Vert^2  
    \leq \cL(z_1) + \frac{1}{\epsilon} \lr  L G^2 \log T \\ &+ \martin \lr \log T(\logtd \frac{G^2}{\epsilon} + \frac{\sigma}{\epsilon}\log^{\frac{1}{2}}(\frac{2T}{\delta_g})) 
    + \kappa^2 \log T (1+\logtsqd)^2 \frac{8}{(1-\beta_1)^2} \frac{\intdim L G^2}{\epsilon^2},
    \end{align*}   
    \end{small}
    where $\delta, \delta_g$, and $\martin$ are the randomness from sketching, sub-Gaussian stochastic noise and martingales respectively, and $J$ is a constant defined in Lemma.~\ref{lemma:non-asym-momentum}. 
    \label{theorem_mild_non_asymptotic}
\end{theorem}

Alike the analysis in the constant learning rate case, we first define auxiliary variables $z_t$
\begin{align*}
    z_t = x_t + \frac{\beta_1}{1-\beta_1}(x_t - x_{t-1}) = \frac{1}{1-\beta_1}x_t - \frac{\beta_1}{1-\beta_1}x_{t-1}.
\end{align*}

Then, the update on $z_t$ can be expressed as
\begin{align*}
    z_{t+1}-z_t &= \frac{1}{1-\beta_1}(x_{t+1}-x_t) - \frac{\beta_1}{1-\beta_1}(x_t-x_{t-1}) \\
    &= \frac{\beta_1}{1-\beta_1}\left(\lr {\hat{V}_{t-1}}^{-1/2}- \lr \hat{V_t}^{-1/2}\right)m_{t-1} - \frac{\lr \eta_t}{C} \hat{V_t}^{-1/2} \sum_{c=1}^C  \sum_{k=1}^K R_t^\top R_t g_{t,k}^c 
\end{align*}

By Taylor expansion, we have
\begin{align*}
    \cL(z_{t+1}) &= \cL(z_t) + \nabla \cL(z_t)^\top (z_{t+1}-z_t) + \frac{1}{2}(z_{t+1}-z_t)^\top \hat H_{\cL} (z_{t+1}-z_t) \\
    &= \cL(z_t) + \nabla \cL(x_t)^\top (z_{t+1}-z_t) +  (\nabla \cL(z_t) - \nabla \cL(x_t))^\top (z_{t+1}-z_t) +\frac{1}{2} (z_{t+1}-z_t)^\top \hat H_{\cL} (z_{t+1}-z_t). 
\end{align*}

Bounding the first-order term
\begin{align*}
&\nabla \cL(x_t)^\top (z_{t+1}-z_t) \\
=& \nabla \cL(x_t)^\top \left(  \frac{\beta_1}{1-\beta_1}\left(\lr {\hat{V}_{t-1}}^{-1/2}- \lr \hat{V_t}^{-1/2}\right)m_{t-1} - \frac{\lr\eta_t}{C} \hat{V_t}^{-1/2} \sum_{c=1}^C  \sum_{k=1}^K R_t^\top R_t g_{t,k}^c \right)\\
\le &  \frac{\beta_1}{1-\beta_1} \Vert \nabla \cL(x_t) \Vert_\infty (\Vert \lr {\hat{V}_{t-1}}^{-1/2}\Vert_{1,1} - \Vert  \lr \hat{V_t}^{-1/2} \Vert_{1,1})\Vert m_{t-1}\Vert_\infty \\
&-  \frac{\eta_t}{C}\nabla \cL(x_t)^\top (\lr \hat{V_t}^{-1/2}-\lr {\hat{V}_{t-1}}^{-1/2})  \sum_{c=1}^C  \sum_{k=1}^K R_t^\top R_t g_{t,k}^c  - \frac{\lr\eta_t}{C}\nabla \cL(x_t)^\top {\hat{V}_{t-1}}^{-1/2}  \sum_{c=1}^C  \sum_{k=1}^K R_t^\top R_t g_{t,k}^c \\
\le & \left(\frac{\beta_1}{1-\beta_1} \Vert m_{t-1}\Vert_\infty  + \frac{\eta_t}{C}\Vert \sum_{c=1}^C  \sum_{k=1}^K R_t^\top R_t g_{t,k}^c  \Vert_\infty\right)  \Vert \nabla \cL(x_t) \Vert_\infty (\Vert \lr {\hat{V}_{t-1}}^{-1/2}\Vert_{1,1} - \Vert  \lr \hat{V_t}^{-1/2} \Vert_{1,1}) \\
& - \frac{\lr\eta_t}{C}\sum_{c=1}^C  \sum_{k=1}^K\nabla \cL(x_t)^\top {\hat{V}_{t-1}}^{-1/2}   R_t^\top R_t g_{t,k}^c.
\end{align*}

The quadratic terms can be written as 
\begin{align*}
    &(\nabla \cL(z_t) - \nabla \cL(x_t))^\top (z_{t+1}-z_t) = (z_t - x_t)^\top \hat{H}_{\cL}(\frac{1}{1-\beta_1}(x_{t+1}-x_t) - \frac{\beta_1}{1-\beta_1}(x_t-x_{t-1})),
\end{align*}
where $\hat{H}_\cL$ is a second-order Taylor remainder. 

To bound the quadratic term, the counterpart of Lemma~\ref{lemma_inner_v} can be stated as
\begin{lemma}
    With learning rate $\eta_t = O(\frac{1}{\sqrt{t+T_0}})$, where $T_0=\lceil \frac{1}{1-\beta_1^2}\rceil$. Denote $J = \frac{1-\beta_1}{\sqrt{T_0+1}}/(\frac{1}{\sqrt{T_0+1}} - \frac{\beta_1}{\sqrt{T_0}}).$ Then with probability $1-\Theta(t\delta)$,
    \begin{align*}
    \vert m_{t-1}^\top h \vert \le (1+\frac{\log^{1.5}(CKd/\delta)}{\sqrt{b}}) JKGH
\end{align*}
\label{lemma:non-asym-momentum}
\end{lemma}
\begin{proof}
 For $t=0$, since $m_0=0$, the inequality holds. Suppose we have for $h \in \mathbb{R}^d$, s.t. $\Vert h\Vert \le H$, with probability $1-\Theta((t-1)\delta)$,
\begin{align*}
    \vert m_{t-1}^\top h \vert \le (1+\frac{\log^{1.5}(CKd/\delta)}{\sqrt{b}}) JKGH
\end{align*}
By the update rule,
    \begin{align*}
\vert m_t^\top h \vert &= \vert (\beta_1 \cdot m_{t-1} + (1 - \beta_1) \cdot \frac{\etat}{C} \sum_{c=1}^C  \sum_{k=1}^K R_t^\top R_t g_{t,k}^c )^\top h\vert \\
& \le \beta_1 \vert m_{t-1}^\top h \vert + \frac{(1-\beta_1)\eta}{C}  \sum_{c=1}^C  \sum_{k=1}^K \vert \langle R_t^\top R_t g_{t,k}^c,h \rangle \vert \\
& \le \beta_1 \vert m_{t-1}^\top h \vert + (1-\beta_1) (1+\frac{\log^{1.5}(CKd/\delta)}{\sqrt{b}})  \eta_t  \sum_{k=1}^K \Vert g_{t,k}^c \Vert_2 \Vert h \Vert_2 \\
& \le (1+\frac{\log^{1.5}(CKd/\delta)}{\sqrt{b}}) \eta_t JKGH,~~w.p.~ 1-\Theta(t \delta).
\end{align*}
\end{proof}

By exactly the same as in Sec.~\ref{sec:proof_lemma_descent_bound}, we can lower bound the scaled gradient term by
\begin{align*}
\nabla \cL(x_t)^\top {\hat{V}_{t-1}}^{-1/2} \nabla \cL(x_t) 
& \ge  \min_i(\kV)_i \sum_{i=1}^d  [\nabla \cL(x_t)]_i^2 \\
&\ge (\sqrt{1+\logstd}\eta_t KG + \epsilon)^{-1} \Vert \nabla \cL (x_t) \Vert^2, ~~w.p.~ 1 -\Theta(d\delta).
\end{align*}

On the martingale of zero-centered noises, we can simply incorporate the learning rate $\eta_t$ into the martingale. Define the random process of sketching noise as
\begin{align*}
    \{Y_t = \sum_{\tau=1}^t\frac{\eta_\tau}{C}\sum_{k=1}^K \nabla \cL(x_\tau)^\top {\hat{V}_{\tau-1}}^{-1/2} (R_\tau^\top R_\tau g_{\tau,k}^c - g_{\tau,k}^c)\}_{t=1}^T
\end{align*}
 as a martingale. The difference of $\vert Y_{t} - Y_{t-1}\vert$ is bounded with high probability
\begin{align*}
    \vert Y_{t} - Y_{t-1}\vert &= \vert \frac{\eta_t}{C} \sum_{c=1}^C \sum_{k=1}^K \nabla \cL(x_t)^\top {\hat{V}_{t-1}}^{-1/2} (R_t^\top R_t g_{t,k}^c - g_{t,k}^c) \vert \\
    &\le \frac{\log^{1.5}(d/\delta)}{\sqrt{b}} \eta_t KG \Vert \hV  \nabla \cL(x_t)\Vert_2, ~~w.p.~ 1-\Theta(CK\delta).
\end{align*}

Then by Azuma's inequality,
\begin{align}
\P(\vert Y_T \vert \ge \martin \sqrt{\sum_{t=1}^T \left( \frac{\log^{1.5}(d/\delta)}{\sqrt{b}} \eta_t KG \Vert \hV  \nabla \cL(x_t)\Vert_2\right)^2} ) = O({\rm exp}(-\Omega(\martin^2))) + T \delta
\end{align}
A similar bound can be achieved for the sub-Gaussian noise in stochastic gradient. Let 
\begin{align*}
    Z_t =  \sum_{\tau=1}^t \frac{\eta_\tau}{C} \sum_{k=1}^K  \nabla \cL(x_\tau)^\top {\hat{V}_{\tau-1}}^{-1/2} ( g_{\tau,k}^c - \nabla \cL^c(x_{t,k}^c)).
\end{align*}
Then
\begin{align*}
    \P(\vert Z_T \vert \ge \martin \sqrt{\sum_{t=1}^T (\frac{\eta_t \sigma}{\epsilon})^2 \log(\frac{2T}{\delta_g}) } ) = O({\rm exp}(-\Omega(\martin^2))) + \delta_g
\end{align*}
Combining the two bounds by union bound completes the proof.

\section{Convergence Without Bounded Gradient Norm Assumption (Proof of Theorem~\ref{theorem_no_grad_norm})}
\label{app:nogradnorm}

We first prove the local client gradient $\cL^c$ is bounded. The client performs stochastic gradient descent $x_{t, k}^c = x_t - \etat \sum_{\tau=1}^k g_{t, \tau}^c$.  Let $\etat = \frac{\eta_0}{\sqrt{K}}$

\begin{lemma}
    Under Assumption~\ref{assume_local_grad_sub},  Let $\etat \le \frac{1}{2L \sqrt{K}}$. The local gradients as of $k \le K$ can be bounded by $\Vert \nabla \cL^c (x_{t,k}^c)\Vert \le \sqrt{2\localnoise^2 \ln \frac{2}{\stocdelta}} + \sqrt{2\localnoise^2 \ln \frac{2}{\stocdelta}+4L \cL^c(x_t)+\localnoise^2}$ with probability $1-K \stocdelta - K \exp(-\localnoise^2/\sigma^2)$.
\end{lemma}
\begin{proof}
\begin{align*}
    &\frac{1}{2L} \Vert \nabla \cL^c(x_{t,k}^c) \Vert^2 \le \cL^c(x_{t,k}^c) \le \cL^c(x_t) + \sum_{k=1}^K \langle \nabla \cL^c(x_t), x_{t,k}^c - x_{t,k-1}^c \rangle + \frac{L}{2}\Vert x_{t,k}^c - x_{t,k-1}^c \Vert^2 \\
    &=  \cL^c(x_t) - \etat \sum_{k=1}^K \langle \nabla \cL^c(x_{t,k}^c), \nabla \cL^c(x_{t,k}^c) + \epsilon_{t,k}^c \rangle + \frac{L}{2}\Vert x_{t,k}^c - x_{t,k-1}^c \Vert^2 \\
    &= \cL^c(x_t) + \etat \sum_{k=1}^K -\Vert \nabla \cL^c(x_{t,k}^c) \Vert^2 - \eta\sum_{\tau=1}^{k-1}\langle \nabla \cL^c(x_{t,\tau}^c) , \epsilon_{t,\tau}^c\rangle + \eta^2 \sum_{\tau=1}^k L (\Vert \nabla \cL^c(x_{t,\tau}^c) \Vert^2 + \Vert \epsilon_{t,\tau}^c \Vert^2)
\end{align*}

Take induction basis $\tau \le k-1$. We have bounded gradient $\Vert \nabla \cL^c (x_{t,\tau}^c)\Vert^2 \le G$ with probability $1- \tau \stocdelta -\tau \exp(-\localnoise^2/\sigma^2)$. The RHS can be bounded with probability $1-k \stocdelta - k \exp(-\localnoise^2/\sigma^2)$ by
\begin{align*}
 &\cL^c(x_t) - \etat \sum_{\tau=1}^{k-1} \langle  \nabla \cL^c(x_{t,\tau}^c), \epsilon_{t,\tau}^c \rangle+ \eta^2 \sum_{\tau=1}^{k-1} L (\Vert \nabla \cL^c(x_{t,\tau}^c) \Vert^2 + \Vert \epsilon_{t,\tau}^c \Vert^2) \\
 \le & \cL^c(x_t) + \frac{\eta_0}{\sqrt{K}} \sqrt{2K G \localnoise^2 \ln \frac{2}{\stocdelta}} + \frac{\eta_0^2 L}{K} K(G + \localnoise^2) \\
 \le & \cL^c(x_t) + \eta_0 \sqrt{2 G \localnoise^2 \ln \frac{2}{\stocdelta}} + \eta_0^2 L (G + \localnoise^2)  \\
 \le & \cL^c(x_t) + \frac{\eta_0}{2} G + \eta_0  \localnoise^2 \ln \frac{2}{\stocdelta} + \eta_0^2 L (G + \localnoise^2)  \le \frac{G}{2L}   
\end{align*}
Let $\eta_0 \le \frac{1}{2L}$, and $G = (\sqrt{2\localnoise^2 \ln \frac{2}{\stocdelta}} + \sqrt{2\localnoise^2 \ln \frac{2}{\stocdelta}+4L \cL^c(x_t)+\localnoise^2})^2$, we have
\begin{align*}
    {\rm RHS} &= \cL^c(x_t) + \frac{\eta_0}{2} G + \eta_0  \localnoise^2 \ln \frac{2}{\stocdelta} + \eta_0^2 L (G + \localnoise^2\ln \frac{2}{\stocdelta}) \\
    & \le \cL^c(x_t) + \frac{1}{4L} G + \frac{1}{2L} \localnoise^2 \ln \frac{2}{\stocdelta} + \frac{1}{4L} (G + \localnoise^2\ln \frac{2}{\stocdelta}) = \frac{G}{2L}
\end{align*}
\end{proof}

Consider the server optimizer
\begin{align*}
    \cL(z_{t+1}) &= \cL(z_t) + \nabla \cL(z_t)^\top (z_{t+1}-z_t) + \frac{1}{2}(z_{t+1}-z_t)^\top \hat H_{\cL} (z_{t+1}-z_t) \\
    &= \cL(z_t) + \nabla \cL(x_t)^\top (z_{t+1}-z_t) +  (\nabla \cL(z_t) - \nabla \cL(x_t))^\top (z_{t+1}-z_t) +\frac{1}{2} (z_{t+1}-z_t)^\top \hat H_{\cL} (z_{t+1}-z_t). 
\end{align*}

\begin{align*}
&\nabla \cL(x_t)^\top (z_{t+1}-z_t) \\
=& \nabla \cL(x_t)^\top \left(  \frac{\beta_1}{1-\beta_1}\left(\lr {\hat{V}_{t-1}}^{-1/2}- \lr \hat{V_t}^{-1/2}\right)m_{t-1} - \frac{\lr\etat}{C} \hat{V_t}^{-1/2} \sum_{c=1}^C  \sum_{k=1}^K R_t^\top R_t g_{t,k}^c \right)\\
=& \frac{\beta_1}{1-\beta_1} \nabla \cL(x_t)^\top \left(\lr {\hat{V}_{t-1}}^{-1/2}- \lr \hat{V_t}^{-1/2}\right)m_{t-1} \\
&- \frac{\lr\etat}{C} \nabla \cL(x_t)^\top\hat{V_t}^{-1/2} \sum_{c=1}^C  \sum_{k=1}^K  \nabla \cL^c(x_t) - \nabla \cL^c(x_t) + \nabla \cL^c(x_{t,k}^c) - \nabla \cL^c(x_{t,k}^c) +g_{t,k}^c  - g_{t,k}^c  + R_t^\top R_t g_{t,k}^c 
\end{align*}

\begin{align*}
    \frac{1}{C} \sum_{c=1}^C \Vert  \nabla \cL^c(x_{t,\tau}^c)\Vert \le & \frac{1}{C} \sum_{c=1}^C \sqrt{2\localnoise^2 \ln \frac{2}{\stocdelta}} + \sqrt{2\localnoise^2 \ln \frac{2}{\stocdelta}+4L \cL^c(x_t)+\localnoise^2}\\
    \le & \frac{1}{C} \sum_{c=1}^C 2\sqrt{L \cL^c(x_t)} + 2\sqrt{2\localnoise^2 \ln \frac{2}{\stocdelta}} + \localnoise \\
    \le & \frac{2\sqrt{L}}{C} \sqrt{C \sum_{c=1}^C \cL^c(x_t)} + 2\sqrt{2\localnoise^2 \ln \frac{2}{\stocdelta}} + \localnoise \\
    = &  2\sqrt{L} \sqrt{\cL(x_t)} + 2\sqrt{2\localnoise^2 \ln \frac{2}{\stocdelta}} + \localnoise 
\end{align*}
where the third inequality follows by Cauchy-Schwarz. On the server side, we consider the induction basis $\frac{1}{2L} \Vert \nabla \cL(x_t)\Vert^2 \le \cL(x_t) \le \frac{G}{2L}$, $w.p.~ 1- t{\rm exp}(-\Omega(\martin^2))- t C\stocdelta - tCK\exp(-\localnoise^2/\sigma^2)$ holds for $t \le T-1$. The following inequality holds with probability $1-K\stocdelta - CK \exp(-\localnoise^2/\sigma^2)$,

\begin{align*}
    &\frac{\lr\etat}{C} \nabla \cL(x_t)^\top\hat{V_t}^{-1/2} \sum_{c=1}^C  \sum_{k=1}^K  - \nabla \cL^c(x_t) + \nabla \cL^c(x_{t,k}^c) \\
    \le & \frac{\lr\etat}{C} \Vert \nabla \cL(x_t)\Vert \Vert \hat{V_t}^{-1/2} \Vert \sum_{c=1}^C  \sum_{k=1}^K  \eta L \Vert \sum_{\tau=1}^{k-1} g_{t,\tau}^c\Vert \\
    \le & \frac{\lr\etat^2 L}{C} \Vert \nabla \cL(x_t)\Vert \Vert \hat{V_t}^{-1/2} \Vert \sum_{c=1}^C  \sum_{k=1}^K  \Vert \sum_{\tau=1}^{k-1} g_{t,\tau}^c\Vert \\
    \le & \frac{\lr\etat^2 L}{\epsilon C} \Vert \nabla \cL(x_t)\Vert \sum_{c=1}^C  \sum_{k=1}^K  \sum_{\tau=1}^{k-1} \Vert  \nabla \cL^c(x_{t,\tau}^c)\Vert  + \localnoise \\
    \le & \frac{\lr\etat^2 K^2 L}{\epsilon} \Vert \nabla \cL(x_t)\Vert (\localgradbnd)  +  \frac{\lr\etat^2 K^2 L}{\epsilon} \Vert \nabla \cL(x_t)\Vert \localnoise \\
    \le & \frac{\sqrt{2}\lr\etat^2 K^2 L^2}{\epsilon} \cL(x_t) + \frac{2\lr\etat^2 K^2 L}{\epsilon} \Vert \nabla \cL(x_t)\Vert \localnoise (1+ \sqrt{2\ln \frac{2}{\stocdelta}})
\end{align*}

And for the difference term, applying Lemma~\ref{lemma_gaussian} yields
\begin{align*}
    &\frac{\etat}{C}\nabla \cL(x_t)^\top (\lr \hat{V_t}^{-1/2}-\lr {\hat{V}_{t-1}}^{-1/2})  \sum_{c=1}^C  \sum_{k=1}^K R_t^\top R_t g_{t,k}^c \\
    \le& \frac{\etat \lr}{C}(1+\logtd) \Vert \nabla \cL(x_t) \Vert \Vert \hat{V_t}^{-1/2}-{\hat{V}_{t-1}}^{-1/2} \Vert_2 \sum_{c=1}^C  \sum_{k=1}^K\Vert g_{t,k}^c \Vert
\end{align*}
Denote $[\cdot]_i$ as the $i$-th element of a vector. The $l2$-norm
\begin{align*}
    \Vert \hat{V_t}^{-1/2}-{\hat{V}_{t-1}}^{-1/2} \Vert_2 & = \max_i \frac{1}{\sqrt{\hat{v}_{t-1, i}}+\epsilon} - \frac{1}{\sqrt{\hat{v}_{t, i}}+\epsilon} = \max_i \frac{\sqrt{\hat{v}_{t, i}} - \sqrt{\hat{v}_{t-1, i}}}{(\sqrt{\hat{v}_{t-1, i}}+\epsilon)(\sqrt{\hat{v}_{t, i}}+\epsilon)} \\
    & = \max_i \frac{\hat{v}_{t, i} - \hat{v}_{t-1, i}}{(\sqrt{\hat{v}_{t-1, i}}+\epsilon)(\sqrt{\hat{v}_{t, i}}+\epsilon)(\sqrt{\hat{v}_{t, i}} + \sqrt{\hat{v}_{t-1, i}})} 
\end{align*}
By definition, $\hat{v}_{t} = \max(\hat{v}_{t-1}, v_t)$. If $\hat{v}_{t, i} = \hat{v}_{t-1, i}$, the RHS is 0. Otherwise, $\hat{v}_{t,i} = v_{t,i}$.
\begin{align*}
     \Vert \hat{V_t}^{-1/2}-{\hat{V}_{t-1}}^{-1/2} \Vert_2 &\le \max_i \frac{v_{t,i} - v_{t-1,i}}{(\sqrt{\hat{v}_{t-1, i}}+\epsilon)(\sqrt{\hat{v}_{t, i}}+\epsilon)(\sqrt{\hat{v}_{t, i}} + \sqrt{\hat{v}_{t-1, i}})} \\
     &\le \max_i \frac{(1-\beta_2)(\barv_{t,i} - v_{t-1,i})}{\epsilon^2 \sqrt{(1-\beta_2)\barv_{t,i}}} \\
     &\le \max_i \frac{\sqrt{1-\beta_2}}{\epsilon^2} \sqrt{\barv_{t,i}} \\
     &= \frac{\sqrt{1-\beta_2}}{\epsilon^2} \max_i \sqrt{\frac{\etat^2}{C}\sum_{c=1}^C [(\sum_{k=1}^K R_t^\top R_t g_{t,k}^c)^2]_i} \\
     &\le  \frac{\etat \sqrt{2(1-\beta_2)}}{\epsilon^2} \sqrt{1+\logdsqt}(\sqrt{2G} + 2 \localnoise (1+ \sqrt{2\ln \frac{2CK}{\stocdelta}})).
\end{align*}
The first inequality is from $\hat{v}_{t-1,i} \ge v_{t-1,i}$. The second inequality comes from $\hat{v}_{t,i} \ge v_{t,i} \ge (1-\beta_2)\barv_{t,i}$. The last inequality follows from applying Lemma~\ref{lemma_gaussian} to each dimension of $g_{t,k}^c$. Plugging into the bound for the difference term
\begin{align*}
&\frac{\etat}{C}\nabla \cL(x_t)^\top (\lr \hat{V_t}^{-1/2}-\lr {\hat{V}_{t-1}}^{-1/2})  \sum_{c=1}^C  \sum_{k=1}^K R_t^\top R_t g_{t,k}^c \\
\le & \frac{\etat^2 \lr  \sqrt{2(1-\beta_2)}}{\epsilon^2} (1+\logdsqt)^{3/2} \sqrt{G}(\sqrt{2G} + 2 \localnoise (1+ \sqrt{2\ln \frac{2CK}{\stocdelta}}))^2
\end{align*}

Consider the sketching noise term. Since the noise is zero-centered, we view the random process of 
\begin{align*}
    \{Y_t = \sum_{\tau=1}^t\sumv{C}\sum_{k=1}^K \nabla \cL(x_\tau)^\top {\hat{V}_{\tau-1}}^{-1/2} (R_\tau^\top R_\tau g_{\tau,k}^c - g_{\tau,k}^c)\}_{t=1}^{T-1}
\end{align*}
as a martingale. The difference of $\vert Y_{t+1} - Y_t\vert$ is bounded with high probability
\begin{align*}
    \vert Y_{t+1} - Y_t\vert \le & \frac{1}{C}\sum_{c=1}^C\sum_{k=1}^K  \vert  \nabla \cL(x_t)^\top {\hat{V}_{t-1}}^{-1/2} (R_t^\top R_t g_{t,k}^c - g_{t,k}^c) \vert \le \sum_{k=1}^K\frac{\log^{1.5}(CKd/\delta)}{\sqrt{b}} \Vert g_{t,k}^c \Vert \Vert {\hat{V}_{t-1}}^{-1/2}\nabla \cL(x_t)\Vert_2 \\
    \le & \frac{\log^{1.5}(CKd/\delta)}{\sqrt{b}}\frac{K}{\epsilon}\left( 2\sqrt{2} L \cL(x_t) + 2 \Vert \nabla \cL(x_t)\Vert \localnoise(1+ \sqrt{2\ln \frac{2CK}{\stocdelta}}) \right)
\end{align*}
Then by Azuma's inequality, with probability at least $1-T{\rm exp}(-\Omega(\martin^2)) - T\stocdelta-TCK\exp(-\localnoise^2/\sigma^2)$
\begin{align*}
\vert Y_T \vert \le & \martin  \frac{\log^{1.5}(CKTd/\delta)}{\sqrt{b}}\frac{K}{\epsilon} \left( \sum_{t=1}^T (2\sqrt{2} L\cL(x_t) + 2\Vert \nabla \cL(x_t)\Vert \localnoise (1+ \sqrt{2\ln \frac{2CK}{\stocdelta}}))^2 \right)^{1/2} \\
\le & \martin  \frac{\log^{1.5}(CKTd/\delta)}{\sqrt{b}}\frac{K}{\epsilon} \sqrt{T} (\sqrt{2} G + 2\sqrt{G}  \localnoise (1+ \sqrt{2\ln \frac{2CK}{\stocdelta}}))
\end{align*}
where the second inequality follows from the induction basis.

We also consider the product term $\vert (\hV m_{t})^\top v_i\vert$.
\begin{lemma}
\label{lemma_inner_nograd}
With probability $1-\Theta(t \delta)$, for eigenvector $v_i$ of the Hessian matrix, $\vert (\hV m_{t})^\top v_i \vert \le (1+\frac{\log^{1.5}(CKd/\delta)}{\sqrt{b}}) \eta K G / \epsilon$.
\end{lemma}

\begin{proof}
We can prove by induction. For $t=0$, since $m_0=0$, the inequality holds. By the induction basis, $\Vert g_{t,k}^c$ has a uniform upper bound.  Suppose we have for $h \in \mathbb{R}^d$, s.t. $\Vert h\Vert \le H$, with probability $1-\Theta((t-1)\delta)$,
\begin{align*}
    \vert m_{t-1}^\top h \vert \le (1+\frac{\log^{1.5}(CKd/\delta)}{\sqrt{b}}) \eta KH (\sqrt{2G}  + 2\localnoise (1+ \sqrt{2\ln \frac{2CK}{\stocdelta}}))
\end{align*}
Then by the update rule,
\begin{align*}
\vert m_t^\top h \vert &= \vert (\beta_1 \cdot m_{t-1} + (1 - \beta_1) \cdot \frac{\etat}{C} \sum_{c=1}^C  \sum_{k=1}^K R_t^\top R_t g_{t,k}^c )^\top h\vert \\
& \le \beta_1 \vert m_{t-1}^\top h \vert + \frac{(1-\beta_1)\eta}{C}  \sum_{c=1}^C  \sum_{k=1}^K \vert \langle R_t^\top R_t g_{t,k}^c,h \rangle \vert \\
& \le \beta_1 \vert m_{t-1}^\top h \vert + (1-\beta_1) (1+\frac{\log^{1.5}(CKd/\delta)}{\sqrt{b}})  \eta  \frac{1}{C} \sum_{c=1}^C \sum_{k=1}^K \Vert g_{t,k}^c \Vert_2 \Vert h \Vert_2 \\
& \le (1+\frac{\log^{1.5}(CKd/\delta)}{\sqrt{b}}) \eta KH (\sqrt{2G} + 2  \localnoise (1+ \sqrt{2\ln \frac{2CK}{\stocdelta}})) ,~~w.p.~ 1-\Theta(t \delta).
\end{align*}
Let $h=\hV v_i$. Then $\Vert h\Vert_2 \le 1/\epsilon$. We have
\begin{align*}
\vert (\hV m_{t})^\top v_i \vert &\le (1+\frac{\log^{1.5}(CKd/\delta)}{\sqrt{b}}) \eta K (\sqrt{2G}  + 2\localnoise (1+ \sqrt{2\ln \frac{2CK}{\stocdelta}})) / \epsilon
\end{align*}
\end{proof}

Then we consider the quadratic term, with probability $1-t \delta - tCK \exp(-\localnoise^2/\sigma^2)$
\begin{align*}
    &(\nabla \cL(z_t) - \nabla \cL(x_t))^\top (z_{t+1}-z_t) \\
    \le & \lr^2 \frac{\beta_1}{(1-\beta_1)^2}(\kV m_{t-1})^\top \hat{H}_{\cL} (\hV m_t) - \lr^2 \frac{\beta_1^2}{(1-\beta_1)^2} (\kV m_{t-1})^\top \hat{H}_{\cL} (\kV m_{t-1}), \\
    =& \lr^2 \frac{\beta_1}{(1-\beta_1)^2}\sum_{i=1}^d \lambda_i  (\kV m_{t-1})^\top (v_i v_i^\top)\hV m_t - \lr^2 \frac{\beta_1^2}{(1-\beta_1)^2} \sum_{i=1}^d \lambda_i  (\kV m_{t-1})^\top (v_i v_i^\top) \kV m_{t-1} \\
    \le & \lr^2 \frac{\beta_1}{(1-\beta_1)^2}\sum_{i=1}^d \vert \lambda_i\vert  \vert(\kV m_{t-1})^\top v_i\vert \vert(\hV m_t)^\top v_i \vert + \lr^2 \frac{\beta_1^2}{(1-\beta_1)^2} \sum_{i=1}^d \vert\lambda_i\vert  \vert (\kV m_{t-1})^\top v_i \vert^2 \\
    \le &  \lr^2 \frac{2}{(1-\beta_1)^2} \intdim L (1+\frac{\log^{1.5}(CKd^2/\delta)}{\sqrt{b}})^2 \eta^2 K^2 (\sqrt{2G}  + 2\localnoise (1+ \sqrt{2\ln \frac{2CK}{\stocdelta}}))^2 / \epsilon^2 \\
    \le &  \lr^2 \frac{8}{(1-\beta_1)^2} \intdim L (1+\frac{\log^{1.5}(CKd^2/\delta)}{\sqrt{b}})^2 \eta^2 K^2 (G  +2\localnoise^2 (1+ \sqrt{2\ln \frac{2CK}{\stocdelta}})^2) / \epsilon^2
\end{align*}
where the last but one inequality is by $\beta_1 \le 1$ and Lemma.~\ref{lemma_inner_nograd}.

Putting all these things together, with probability $ 1- T{\rm exp}(-\Omega(\martin^2))- T C\stocdelta - TCK\exp(-\localnoise^2/\sigma^2) - T\delta$
\begin{align*}
\cL(x_T) =& \cL(z_{T}) + \frac{\beta_1}{1-\beta_1}\langle \nabla \cL(z_T), \lr \hat{V}_{t}^{-1/2}m_t\rangle + \frac{\lr^2}{2}\frac{\beta_1^2}{(1-\beta_1)^2} (\hat{V}_{t}^{-1/2}m_t)^\top H_\cL (\hat{V}_{t}^{-1/2}m_t) \\
    =& \cL(z_1) + \sum_{t=1}^{T-1} \nabla \cL(x_t)^\top (z_{t+1}-z_t) +  (\nabla \cL(z_t) - \nabla \cL(x_t))^\top (z_{t+1}-z_t) +\frac{1}{2} (z_{t+1}-z_t)^\top \hat H_{\cL} (z_{t+1}-z_t) \\
    &  + \frac{\beta_1}{1-\beta_1}\langle \nabla \cL(z_T), \lr \hat{V}_{t}^{-1/2}m_t\rangle + \frac{\lr^2}{2}\frac{\beta_1^2}{(1-\beta_1)^2} (\hat{V}_{t}^{-1/2}m_t)^\top H_\cL (\hat{V}_{t}^{-1/2}m_t) \\
    \le& 2 \cL(z_1) + \frac{8\eta\beta_1}{(1-\beta_1)\epsilon} + \sum_{t=1}^{T-1} \frac{2\sqrt{2}\lr\etat^2 K^2 LG}{\epsilon}  + \frac{4\lr\etat^2 K^2 L\sqrt{G}}{\epsilon} \localnoise(1+ \sqrt{2\ln \frac{2CK}{\stocdelta}})\\
    &+ \sum_{t=1}^{T-1}\frac{2\etat^2 \lr  \sqrt{2(1-\beta_2)}}{\epsilon^2} (1+\logdsqt)^{3/2} \sqrt{G}(\sqrt{2G} + 2 \localnoise (1+ \sqrt{2\ln \frac{2CK}{\stocdelta}}))^2\\
    &+ 2\lr \etat \martin  \frac{\log^{1.5}(CKTd/\delta)}{\sqrt{b}}\frac{K}{\epsilon} \sqrt{T} (\sqrt{2} G + 2\sqrt{G}  \localnoise (1+ \sqrt{2\ln \frac{2CK}{\stocdelta}})) + 2\lr \etat \martin \sqrt{T} \frac{K\sqrt{G}}{\epsilon} \localnoise \\
    &+ \sum_{t=1}^{T-1} \lr^2 \frac{16}{(1-\beta_1)^2} \intdim L (1+\frac{\log^{1.5}(CKd^2/\delta)}{\sqrt{b}})^2 \etat^2 K^2 (G  +2\localnoise^2 (1+ \sqrt{2\ln \frac{2CK}{\stocdelta}})^2) / \epsilon^2\\
    &+ \sum_{t=1}^{T-1} \left(\frac{1+\beta_1}{1-\beta_1} \right)^2 (1+\frac{\log^{1.5}(CKd^2/\delta)}{\sqrt{b}})^2 2\lr^2\etat^2 K^2 \intdim L (G  +2\localnoise^2 (1+ \sqrt{2\ln \frac{2CK}{\stocdelta}})^2)/ \epsilon^2 \\
    \le& 2 \cL(z_1) + \frac{8\eta\beta_1}{(1-\beta_1)\epsilon}  + \frac{\lr\eta_0^2 K^2 L\sqrt{G}}{\epsilon} 4\sqrt{2G} 
    + \frac{\eta_0^2 \lr  \sqrt{2(1-\beta_2)}}{\epsilon^2} (1+\logdsqt)^{3/2} 16G^{3/2} \\
    &+ \lr \eta_0 \martin  \frac{\log^{1.5}(CKTd/\delta)}{\sqrt{b}}\frac{K}{\epsilon} 4\sqrt{2} G+ 2\lr \eta_0 \martin \frac{K\sqrt{G}}{\epsilon} \localnoise \\
    &+ \lr^2 \frac{8+(1+\beta_1)^2}{(1-\beta_1)^2} \intdim L (1+\frac{\log^{1.5}(CKd^2/\delta)}{\sqrt{b}})^2 4\eta_0^2 K^2 G  / \epsilon^2 \\
    \le&  2 \cL(z_1) + \frac{8\eta\beta_1}{(1-\beta_1)\epsilon}+ \frac{\eta_0^2 K^2 L}{\epsilon} 4\sqrt{2G} 
    + \frac{\eta_0^2  \sqrt{2(1-\beta_2)}}{\epsilon^2} (1+\logdsqt)^{3/2} 16G \\
    &+ \eta_0 \martin  \frac{\log^{1.5}(CKTd/\delta)}{\sqrt{b}}\frac{K}{\epsilon} 4\sqrt{2G} + 2\eta_0 \martin \frac{K}{\epsilon}\localnoise \\
    &+ \frac{8+(1+\beta_1)^2}{(1-\beta_1)^2} \intdim L (1+\frac{\log^{1.5}(CKTd^2/\delta)}{\sqrt{b}})^2 4\eta_0^2 K^2  / \epsilon^2 \
\end{align*}
where the second inequality holds by $\sqrt{2G} \ge 2 \localnoise (1+ \sqrt{2\ln \frac{2CK}{\stocdelta}})$, the third inequality holds by $\lr \le \frac{1}{\sqrt{G}}$. Let 
\begin{align}
   &\eta_0 \le \frac{\epsilon}{2\sqrt{L}} \min\{\frac{1}{3}, \frac{1-\beta_1}{2\beta_1\sqrt{L}}\}  (1+\frac{\log^{1.5}(CKTd^2/\delta)}{\sqrt{b}})^{-1} \notag\\
   & G \ge \max \{ 2 \localnoise^2 (1+ \sqrt{2\ln \frac{2CK}{\stocdelta}})^2, 512( \frac{\eta_0^2 K^2 L^2}{\epsilon} +  \frac{\log^{1.5}(CKTd/\delta)}{\sqrt{b}}\frac{\eta_0 \martin K}{\epsilon})^2 + 32L(\cL(z_1)+  \frac{4\eta\beta_1}{(1-\beta_1)\epsilon}+ \frac{\eta_0 \martin K \localnoise}{\epsilon} \\
   &~~~~~~~~~~~~~~~+ \frac{8+(1+\beta_1)^2}{(1-\beta_1)^2}  (1+\frac{\log^{1.5}(CKTd^2/\delta)}{\sqrt{b}})^2 \frac{2\eta_0^2 K^2 \intdim L}{\epsilon^2} )\}
   \label{eq:g_bound}
\end{align}
suffice to yield ${\rm RHS} \le \frac{G}{2L}$.

Furthermore, the dropped positive terms regarding the gradient norm is $\sum_{t=1}^T \lr \etat  K \nabla\cL(x_t)^\top \hat{V}_t^{-1/2} \nabla\cL(x_t) \ge \lr\eta_0 \sqrt{T}L \left(\sqrt{1+\logtdsq} \eta K (\sqrt{2G}+2 \localnoise(1+ \sqrt{2\ln \frac{2}{\stocdelta}})) + \epsilon \right)^{-1}\Vert \nabla\cL(x_t) \Vert^2 $. Rearranging the terms yields the convergence result.

Finally, we give the full forms of $\{\cM_i\}_{i=1}^7$,
\begin{align*}
    &\cM_1 := 2\sqrt{2}\martin  \frac{\log^{1.5}(CKTd/\delta)}{\sqrt{b}}\frac{K}{\epsilon}  \\
    &\cM_2 := 4\sqrt{2}\martin  \frac{\log^{1.5}(CKTd/\delta)}{\sqrt{b}}\frac{K}{\epsilon} 2\sqrt{G}  \localnoise (1+ \sqrt{2\ln \frac{2CK}{\stocdelta}}) +  2\martin \frac{K}{\epsilon} \localnoise \\
    &\cM_3 := 2 \cL(z_1) + \frac{8\eta\beta_1}{(1-\beta_1)\epsilon} \\
    &\cM_4 := \frac{4 \sqrt{2(1-\beta_2)}}{\epsilon^2} (1+\logdsqt)^{3/2} \\
    &\cM_5 := \frac{2\sqrt{2} K^2 L}{\epsilon}\\
    &\cM_6 := \frac{4 K^2 L}{\epsilon} \localnoise(1+ \sqrt{2\ln \frac{2CK}{\stocdelta}}) + \frac{4\etat^2 \lr  \sqrt{2(1-\beta_2)}}{\epsilon^2} (1+\logdsqt)^{3/2}\ln \frac{2CK}{\stocdelta}\\
    &\cM_7 := 4 \frac{8+(1+\beta_1)^2}{(1-\beta_1)^2} \intdim (1+\frac{\log^{1.5}(CKd^2/\delta)}{\sqrt{b}})^2 K^2  / \epsilon^2 \\
\end{align*}

\section{Experimental Details and Additional Results}
\label{app:expts}
\begin{figure}[]
\vspace*{-2mm}
    \centering
    \includegraphics[scale=0.2]{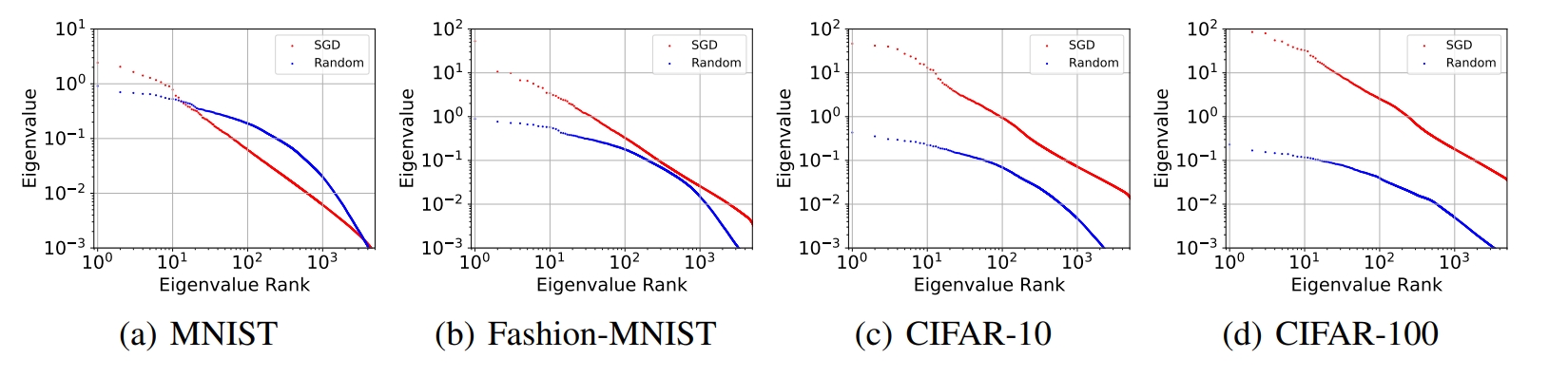}
    \caption{The power-law structure of the Hessian spectrum on LeNet. Quoted from Fig.1~\cite{xie2022power}.}
    \label{fig:li-eigenspectrum}
\vspace*{-3mm}
\end{figure}

\begin{figure}[]
\vspace*{-2mm}
    \centering
    \includegraphics[scale=0.35]{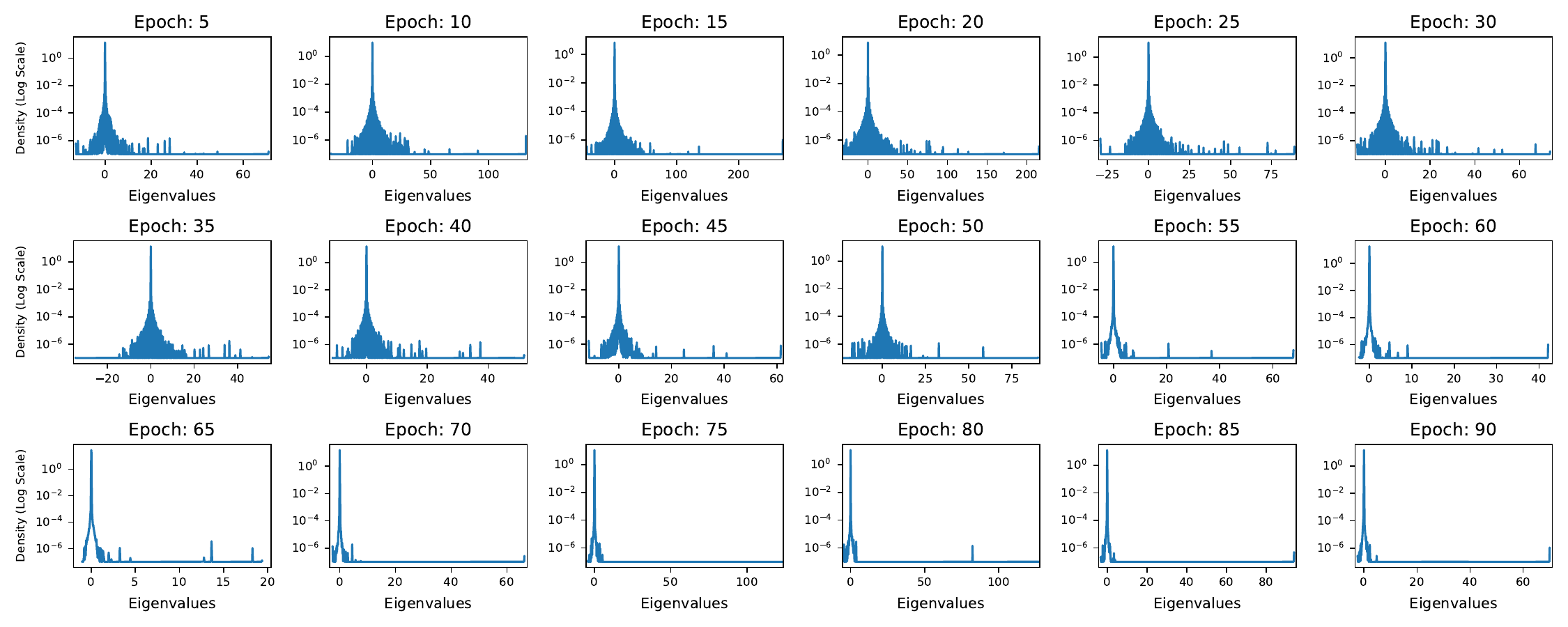}
    \vspace*{-3mm}
    \caption{Eigenspectrum density every 5 epochs. The model is ViT-Small and trained on CIFAR10. The majority of eigenvalues concentrates near 0 and the density enjoys a super fast decay with the absolute values of eigenvalues, indicating a summable eigenspectra.}
    \label{fig:vit-small-eigenspectrum}
\vspace*{-3mm}
\end{figure}

\begin{figure}[]
    \begin{subfigure}{0.5\linewidth}
        \centering
        \includegraphics[scale=0.7]{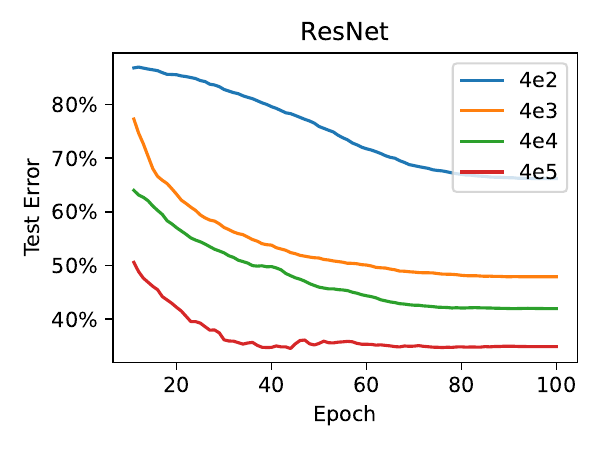}
    \end{subfigure}
    \begin{subfigure}{0.5\linewidth}
        \centering
        \includegraphics[scale = 0.7]{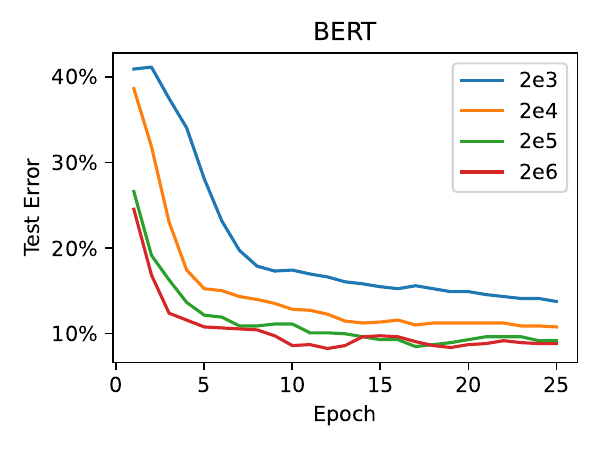}
    \end{subfigure}
    \caption{Comparing the performance of tiny sketch sizes on ResNet and BERT. The experiment settings are the same as in Fig.~\ref{fig:cifar-sketch} and Fig.~\ref{fig:sst-sketch}}.
    \label{fig:sketch_tiny}
\end{figure}

Aside from the experimental configurations described in the main paper, we provide additional details. We use Cross Entropy with label smoothing as the loss function. The parameter for label smoothing is 0.1. We use a cosine learning rate scheduler on the server optimizer, with the minimal learning rate is $1e-5$. Client batch size is 128, and weight decay is $1e-4$. For SGD and SGDm methods, the learning rate is 1.0. For SGDm, the momentum is 0.9. For Adam optimizer, the learning rate is 0.01, and the momentum is 0.9. The learning rates are tuned to achieve the best performance.

Our experiments were conducted on a computing cluster with AMD EPYC 7713 64-Core Processor and NVIDIA A100 Tensor Core GPU. 

To verify Assumption~\ref{assume_hessian}, we plot the full Hessian eigenspectrum throughout the training process in Fig.~\ref{fig:vit-small-eigenspectrum}. We used stochastic lanczos algorithm implemented by the pyHessian library~\cite{yao2020pyhessian} to approximate the distribution of the full eigenspectrum. Our main claim in Assumption~\ref{assume_hessian} is that the Hessian eigenspectrum at an iterate is summable and the sum is independent of the ambient dimension, which can be satisfied by common distributions, like power-laws. We run testing experiments on ViT-small and train on CIFAR-10 dataset, with sketched Adam optimizer. In Fig.~\ref{fig:vit-small-eigenspectrum}, we see the majority of eigenvalues concentrates near 0. The density enjoys a super fast decay with the absolute values of eigenvalues. The decay also holds throughout the training process. This empirical evidence shows the validity of our assumption.

In the main body of the paper, we have achieved 99.9\% compression rate and 99.98\% compression rate for ResNet and BERT respectively. We further include the results on smaller $b$ in Fig.~\ref{fig:sketch_tiny}. In principle, an extremely tiny sketch size (with 400 in vision tasks and 2000 in language tasks) still converges but generates an unfavorable local minima that hardly generalizes.

Additionally, we summarize the theoretical guarantees of the existing approaches in Table~\ref{tab:theory_compare}. From the table, we can see all the comparisons made in the main paper are fair.

\begin{table}[]
    \centering
    \begin{tabular}{|c||c|c|c|} \hline 
        Algorithms & Communication Bits & learning rate & Convergence Rate\\ \hline \hline 
        FetchSGD & $\tilde{O}(1)$ & $O(1/\sqrt{T})$ & $O(1/\sqrt{T}) ~^{(A)}$\\
        CocktailSGD & $O(1)$ & $O(1/(\sqrt{T}+T^{1/3}d^2+d^3))$ & $O(1/\sqrt{T} + d^2/(T)^{2/3})$\\
        CD-Adam & $O(1)$ & $O(1/\sqrt{d})$ & $O(\sqrt{d}/\sqrt{T})$\\
        Onebit-Adam & $O(d)$ & $O(1/\sqrt{T})$ & $O(1/\sqrt{T})$\\
        MARINA & $O(1)$ & $(1+\sqrt{\omega (d-b)/(bC)})^{-1}$ & $O(\sqrt{\frac{\omega}{n}(\frac{d}{b}-1)}/T)~^{(B)}$\\
        Ours & $\tilde{O}(1)$ & $O(1/\sqrt{T})$ & $O(1/\sqrt{T}) ~^{(C)}$ \\ \hline  
    \end{tabular}
    \caption{Comparison on Theoretical Guarantees. We only include the dependence on $d$ and $T$. (A) Needs a heavy-hitter assumption, otherwise deteriorated to $O(T^{1/3})$. (B) The rate is achieved either under deterministic case or use variance reduction methods. $\omega$ is typically $\Theta(d/b)$ when the compressor is RandK or $l_2-$quantization. (C) requires the assumption on the fast-decay Hessian eigenspectrum. Otherwise, the convergence rate can deteriorate to $O(d/\sqrt{T})$ under dimension-independent learning rate. }
    \label{tab:theory_compare}
\end{table}


\end{document}